\documentclass[manuscript,screen,nonacm]{acmart}
\authorsaddresses{}
\usepackage{amsmath,booktabs,array,longtable,placeins,needspace}
\makeatletter
\newcommand{\manuscripttablefont}{\rmfamily\small
  \typeout{FAccT table typography: \f@family/\f@size}}
\makeatother
\AtBeginEnvironment{tabular}{\manuscripttablefont}
\AtBeginEnvironment{longtable}{\manuscripttablefont}
\DeclareCaptionTextFormat{referencechecks}{Historical numerical-reference checks. The checks cover the broader numerical evaluation, including calibration and candidate models, beyond the selected audit histories. Sensitivities are estimates and do not certify error bounds.}
\DeclareCaptionTextFormat{numericaltransitions}{Saved primary-margin decision changes by method. C denotes a conclusive category and I an inconclusive result. Columns count C$\to$I, I$\to$C and changes between conclusive categories.}
\DeclareCaptionTextFormat{basicmainreadable}{TPR audits at $n=1{,}250$ comparing interval form and resample size. Each cell contains twenty cadence histories and 2,000 audits. The margin is 0.5 pp. Coverage averages the twenty history-specific percentages equally. The remaining columns count correct conclusive decisions, false reductions (FR), false increases (FI), false within-margin declarations (FW) and inconclusive results (Inc.) across all audits. Reduced-size basic uses fixed $\gamma=0.75$.}
\DeclareCaptionTextFormat{directionalhighlightsreadable}{Coverage and audit decisions are reported for three controlled configurations, with 2,000 audits per row and a 0.5 pp decision margin. Coverage is expressed as a percentage, while the decision columns give observed counts across all 2,000 audits, with incorrect conclusions classified by the category falsely declared.}
\DeclareCaptionTextFormat{finalcaptionexactbias}{Finite-binomial expectation for the frozen-zero case, in pp. The updating and frozen group-0 rates are 0.5; their group-1 rates are 0.55 and 0.5.}
\DeclareCaptionTextFormat{finalcaptionpanelconditionalsummary}{Conditional coverage and correct conclusive decisions for cadence in the positive-signal interaction cell under subgroup drift. All sizes use the same twenty histories with model changes and 100 audits per history. Posterior coverage concerns TPR; percentile decisions use the 0.5 pp margin. Percentages round exact count fractions half up. History 259 is highlighted retrospectively and has the lowest posterior TPR coverage at all three sizes. Its brackets are pointwise exact 95\% binomial Monte Carlo intervals, not selection-adjusted or population-of-histories intervals.}
\DeclareCaptionTextFormat{finalcaptionmoutnpanel}{Cadence in the selected positive-signal interaction cell under subgroup drift: equal means across all 20 retained histories, each with 100 audits. M denotes the reduced-size basic interval with $\gamma=0.75$; P denotes the saved full-size percentile. Coverage and decision frequencies are percentages; decisions use the 0.5 pp margin.}
\DeclareCaptionTextFormat{finalcaptionreplicationmain}{Combined-drift cadence comparison with disjoint seeds, using twenty selected histories with 100 audits each. P is the saved full-size percentile; M uses fixed $\gamma=0.75$. Frequencies are equal history means (\%); the margin is 0.5 pp. The final column counts lower/equal/higher M coverage on paired evaluable audits. One P TPR history at $n=20{,}000$ has 99 evaluable coverage repeats; its paired comparison also uses 99. Other coverage series retain 100.}
\DeclareCaptionTextFormat{finalcaptiontransitionsmain}{Paired decision transitions in TPR at $n=1{,}250$, margin 0.5 pp. Each block contains the same 2,000 audits of twenty histories. Rows are the earlier method, columns the later method: I = inconclusive, C = correct conclusive, W = incorrect conclusive. All nine cells, including lost conclusions and errors that persist, are retained. These pooled counts describe transitions within each fixed history set.}
\newcolumntype{L}[1]{>{\raggedright\arraybackslash}p{#1}}
\newcommand{\Ex}{\mathbb{E}}
\newcommand{\Prob}{\mathbb{P}}
\newcommand{\reporttable}[1]{\input{editorial/#1}}
\newcommand{\ExactFrozenBias}{1.994861}
\newcommand{\ExactUpdatedBias}{0.041854}
\newcommand{\ExactContrastBias}{-1.953008}

\newcommand{\ExactExpectedContrast}{3.047}

\newcommand{\PairingSame}{0.516}
\newcommand{\PairingOpposite}{1.505}

\newcommand{\NullFalseReductionCount}{84}
\newcommand{\NullFalseReductionPercent}{4.20}
\newcommand{\NullFalseReductionLower}{3.36}
\newcommand{\NullFalseReductionUpper}{5.17}

\newcommand{\MixedFalseReductionCount}{136}
\newcommand{\MixedFalseReductionPercent}{6.80}
\newcommand{\MixedFalseReductionLower}{5.74}
\newcommand{\MixedFalseReductionUpper}{7.99}

\newcommand{\MOutFrozenCoverage}{58.80}
\newcommand{\MOutFrozenCorrect}{94.35}
\newcommand{\MOutFrozenWidth}{2.95}

\newcommand{\MOutMixedFalseReduction}{129}

\newcommand{\MOutHistoryCoverageSmall}{75}
\newcommand{\MOutHistoryCoverageMedium}{89}
\newcommand{\MOutHistoryCoverageLarge}{89}

\newcommand{\MOutNumericalChanges}{1}
\newcommand{\MOutMaximumEndpointShift}{0.354}

\newcommand{\HistoryTPRLowerSmall}{18}
\newcommand{\HistoryTPRMedianSmall}{-15.0}
\newcommand{\HistoryTPRLowerMedium}{17}
\newcommand{\HistoryTPRMedianMedium}{-11.5}
\newcommand{\HistoryTPRLowerLarge}{19}
\newcommand{\HistoryTPRMedianLarge}{-7.5}
\newcommand{\ResolvedProjectionLower}{-14.504}
\newcommand{\ResolvedProjectionUpper}{-1.961}

\newcommand{\DecisionBootstrapWrongConclusive}{3.70}

\newcommand{\DecisionMOutWrongConclusive}{11.19}
\newcommand{\BootstrapHistoryCoverageSmall}{61}
\newcommand{\BootstrapHistoryCoverageMedium}{60}
\newcommand{\BootstrapHistoryCoverageLarge}{76}

\newcommand{\ReplicationTPRLowerSmall}{18}

\newcommand{\ReplicationTPRLowerMedium}{19}

\newcommand{\ReplicationTPRLowerLarge}{18}

\newcommand{\BroadFPRLowerSmall}{148}
\newcommand{\BroadFPREqualSmall}{97}
\newcommand{\BroadFPRHigherSmall}{54}
\newcommand{\BroadFPRLowerMedium}{88}

\newcommand{\BroadFPRHigherMedium}{55}
\newcommand{\BroadFPRLowerLarge}{68}

\newcommand{\BroadFPRHigherLarge}{51}
\newcommand{\BroadTPRLowerSmall}{229}
\newcommand{\BroadTPREqualSmall}{51}
\newcommand{\BroadTPRHigherSmall}{19}
\newcommand{\BroadTPRLowerMedium}{153}

\newcommand{\BroadTPRHigherMedium}{37}
\newcommand{\BroadTPRLowerLarge}{96}

\newcommand{\BroadTPRHigherLarge}{36}

\newcommand{\MarginBroadGroups}{138}
\newcommand{\MarginBroadReversals}{59}

\newcommand{\AggregateSubgroupPBCoverageSE}{0.75}
\newcommand{\AggregateSubgroupPBCoverageLower}{-9.13}
\newcommand{\AggregateSubgroupPBCoverageUpper}{-6.17}
\newcommand{\AggregateSubgroupPBCorrectDelta}{14.90}
\newcommand{\AggregateSubgroupPBCorrectSE}{0.75}

\newcommand{\AggregateSubgroupPBIncorrectDelta}{2.00}
\newcommand{\AggregateSubgroupPBIncorrectSE}{0.31}

\newcommand{\AggregateSubgroupBMErrorAllDelta}{1.05}

\newcommand{\AggregateSubgroupBMErrorAllLower}{0.61}
\newcommand{\AggregateSubgroupBMErrorAllUpper}{1.49}
\newcommand{\AggregateSubgroupBMErrorConclusiveDelta}{0.93}

\newcommand{\AggregateSubgroupBMErrorConclusiveLower}{-0.74}
\newcommand{\AggregateSubgroupBMErrorConclusiveUpper}{2.59}

\hypersetup{pdfauthor={Aaron Ceross and Tingting Zhu},pdftitle={Fairness Auditing After Model Updating: Reliability of Paired Disparity Comparisons}}
\title[Fairness Auditing After Model Updating]{Fairness Auditing After Model Updating: Reliability of Paired Disparity Comparisons}
\author{Aaron Ceross}
\affiliation{\institution{Birmingham Law School, University of Birmingham}\city{Birmingham}\country{United Kingdom}}
\author{Tingting Zhu}
\affiliation{\institution{Department of Engineering Science, University of Oxford}\city{Oxford}\country{United Kingdom}}
\begin{document}
\begin{abstract}
Model updates can change disparities in prediction errors between demographic groups. We examine whether retrospective audits can reliably assess these changes by comparing updated models with their unchanged initial models on the same independent labelled observations, holding completed updating histories fixed. The target is the updated-minus-frozen difference in absolute subgroup gaps, averaged equally across reporting periods and assessed separately for true-positive and false-positive rates. Exploratory controlled simulations and repeated audits of fixed simulated histories assess coverage and decision correctness for nominal 95\% intervals. Across the broader history panel, changing interval construction can raise or lower estimated coverage. In one subgroup-drift setting, 2,000 true-positive-rate audits evaluated twenty selected histories updated at fixed intervals, using 1,250 observations per reporting period. Replacing percentile bootstrap intervals with full-size basic intervals reduced estimated coverage from 92.20\% to 84.55\%. At an illustrative margin of 0.5 percentage points, correct conclusive decisions increased from 78 to 376, while incorrect decisions increased from 3 to 43, including 13 false reductions. The constructions used identical observations, estimates and bootstrap draws and had equal widths. Other controlled examples frequently excluded the population change while correctly indicating an increase. More conclusive audit results therefore need not provide more reliable evidence. Claims that updating reduced disparity require an inference procedure justified for the target and sampling design, an explicit decision margin, and group-specific performance reporting.
\end{abstract}
\maketitle
\section{Introduction}
Updating a predictive model to maintain overall performance can also change disparities in prediction errors between demographic groups. In a longitudinal clinical evaluation, \citet{davis2025fairnessdrift} found that updating could narrow, widen or leave subgroup performance gaps unchanged. Assessing an updating programme therefore requires evidence about how disparities changed relative to retaining the initial model. The reliability of this evidence has consequences for model oversight. A false declaration of disparity reduction could give an organisation reassurance about an updating programme even when the population gap has widened, potentially discouraging further investigation. The EU AI Act places ongoing evaluation within the governance of high-risk AI systems through its provisions for lifecycle risk management and systematic post-market monitoring, including consideration of risks to fundamental rights \citep[Articles~9(2) and~72(1)--(2)]{eu2024aiact}. Claims that updating reduced disparity may also contribute to the context-specific assurance arguments described by \citet{sabuncuoglu2025evidence}, through which statistical findings are considered alongside other evidence about a system and its use.

An audit estimates disparities from a finite sample, so an apparent reduction may reflect sampling variability rather than a population change. Uncertainty intervals are intended to account for this variability, but their reliability depends on the quantity being estimated and the sampling design. \citet{lum2022debiasing} show that sampling variability can systematically inflate disparity summaries and that correcting an estimator's bias does not necessarily calibrate its bootstrap interval. For a comparison of absolute subgroup gaps, the biases in the two estimated gaps need not cancel. Averaging their differences across reporting periods can then reduce variability while retaining bias, concentrating estimates around a value displaced from the population change.

These difficulties connect fairness measurement to established problems in statistical inference. \citet{fang2019inference} characterise the conditions under which the ordinary bootstrap consistently approximates uncertainty for directionally differentiable functions, providing a theoretical basis for examining nonsmooth summaries such as absolute gaps. In a related fairness setting, \citet{wastvedt2024intersectional} develop rescaled inference for aggregates of absolute counterfactual error-rate differences and show that intervals of equal length can have different coverage. For an audit comparing updated and unchanged models, these concerns raise a practical question about how often an interval procedure supports a correct conclusion about disparity change.

\begin{figure}[htbp]\centering
\includegraphics[width=.95\linewidth]{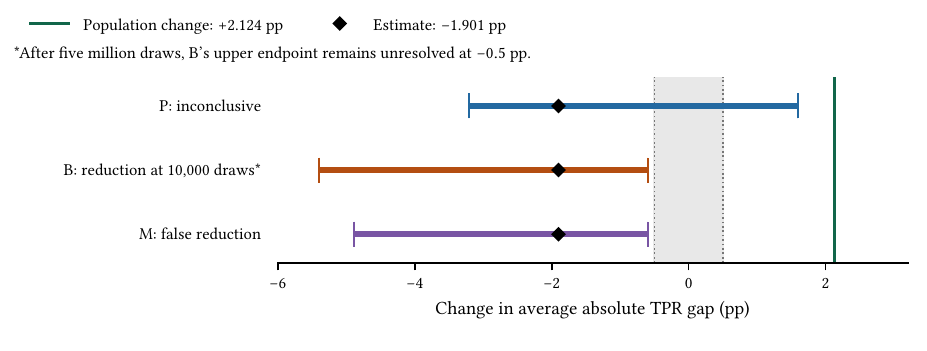}
\caption{Original 10,000-draw intervals for one retrospectively selected audit (history 31, repeat 93, subgroup-drift TPR, $n=1{,}250$ per window). At the 0.5 pp margin, percentile (P) is inconclusive, while full-size basic (B) and reduced-size basic (M, $\gamma=0.75$) declare reduction despite a population increase of $+2.124$ pp. The estimate is $-1.901$ pp, and P and B have identical widths. A separate five-million-draw check leaves B's upper endpoint numerically unresolved at $-0.5$ pp, while M's remains below the boundary. Appendix~\ref{app:figure1-stability} reports the precision assessment.}
\Description{Three saved 10,000-draw intervals share an estimate of minus 1.901 percentage points. Percentile is inconclusive and crosses both margins. Both saved basic intervals lie below minus 0.5 percentage points and falsely declare reduction. The population change is plus 2.124 percentage points, outside all three intervals. An asterisk on the full-size basic label points to the note that the five-million-draw check leaves its verdict unresolved at minus 0.5 percentage points. The reduced-size basic upper endpoint remains below that boundary in the check.}
\label{fig:disagreement}
\end{figure}

We investigate this question by auditing completed updating histories, comparing the updated model in each reporting period with the unchanged initial model on the same independent labelled observations. These observations are reserved from training and every update decision. We subtract the initial model's absolute subgroup gap from the updated model's absolute subgroup gap and average these differences equally across periods, assessing true-positive and false-positive rates separately. The fitted models and realised updating decisions remain fixed across repeated audits, allowing us to examine uncertainty arising from audit sampling. Our exploratory controlled simulations and repeated audits of fixed simulated histories assess both interval coverage and the conclusions supported relative to a stated decision margin. Coverage measures how often an interval contains the population change, while a conclusive audit classifies that change as a reduction, an increase or a difference within the margin. These outcomes can diverge. Some evaluated procedures frequently exclude the population change while correctly indicating an increase. Other comparisons show that changing interval construction on the same observations and bootstrap draws can yield more correct conclusive decisions together with more incorrect decisions, including false declarations of disparity reduction, even when interval widths are unchanged. Figure~\ref{fig:disagreement} illustrates this disagreement for one selected audit, with the numerical stability of its displayed endpoints assessed in Appendix~\ref{app:figure1-stability}.

By tracking the same audits across procedures, we examine which conclusions change and whether those changes agree with the population comparison. The findings show why the frequency of conclusive results and the width of uncertainty intervals provide incomplete evidence of audit reliability. A claim that updating reduced disparity requires an inference procedure justified for the target and sampling design, together with an explicit decision margin. Its interpretation also requires group-specific performance reporting, since a smaller absolute gap can result from worsening performance for one group and does not by itself establish broader fairness.

\FloatBarrier
\section{The paired disparity comparison}
\label{sec:target}
We compare the updated model with the unchanged initial model in each reporting period's population, holding the completed updating history fixed. That history comprises the fitted classifiers and realised update decisions over a prespecified horizon, so the uncertainty examined here arises from sampling independent audit observations.

For reporting periods $t=0,\ldots,T-1$, let $f_{u,t}$ denote the updated classifier and $f_f$ the frozen initial classifier, with $f_{f,t}=f_f$. These are deterministic prediction functions that include preprocessing and the prediction cutoff. The updated classifier may differ across periods, while each period's fitted classifier remains fixed across repeated audits. The frozen classifier's prediction function remains unchanged throughout, although its rates can vary as the population changes. We assess true-positive and false-positive rates separately. Let $A\in\{0,1\}$ denote group, with $Y=y$ selecting TPR when $y=1$ and FPR when $y=0$. In the common population $\Pr_t$ for reporting period $t$, define the group-specific rates and signed subgroup gaps as
\begin{equation}
r_{k,ta}=\Pr_t\{f_{k,t}(X,A)=1\mid A=a,Y=y\},\qquad
g_{k,t}=r_{k,t1}-r_{k,t0},\qquad k\in\{u,f\}.
\end{equation}
Each conditioning event must have positive population probability. With rates and signed gaps expressed as proportions, the target in percentage points (pp) is
\begin{equation}
D=\frac{100}{T}\sum_{t=0}^{T-1}\bigl(|g_{u,t}|-|g_{f,t}|\bigr).
\label{eq:target}
\end{equation}
The absolute values measure the size of each model's subgroup gap regardless of which group has the higher rate. Equation~\eqref{eq:target} subtracts the frozen model's absolute gap from the updated model's absolute gap within each period, then averages these differences over the horizon. Each period receives weight $1/T$, regardless of audit sample size or whether an update occurred. Reductions in some periods can therefore offset increases in others, and $D$ need not have the same sign as the contrast in the final period. Periods in which the prediction functions are established to be identical contribute zero while remaining in $T$. This identity must be established from model records, since agreement on an audit sample alone does not establish that the prediction functions coincide. A negative $D$ indicates lower average absolute disparity under updating, but interpreting this reduction requires examining what happened to each group. Updating could, for example, narrow the absolute TPR gap by lowering TPR in the group with the higher rate while leaving the other group's TPR unchanged. We therefore also examine each group's temporal rate change, $\Delta_a=100T^{-1}\sum_t(r_{u,ta}-r_{f,ta})$. Higher TPR and lower FPR are favourable in their respective error dimensions, with their relative importance and consequences for affected people depending on the application.

To estimate the aforementioned quantities, both classifiers score the same fresh labelled observations within each reporting period. Each observation therefore produces a pair of predictions, whose dependence is retained in the analysis. Conditional on the fixed history, audit observations are independent within and between strata defined by reporting period, group and outcome. They are reserved from training and every update decision, including later decisions, so that the audit evaluates the completed history using observations that played no part in determining it.

For a stratum with group/outcome denominator $n_{ta}>0$, the four possible prediction pairs have counts
\begin{equation}
\mathbf C_{ta}=(C_{00},C_{01},C_{10},C_{11})_{ta}
\sim\operatorname{Multinomial}(n_{ta},\mathbf q_{ta}),
\end{equation}
where the first cell index is the updated prediction and the second is the frozen prediction. The vector $\mathbf q_{ta}$ contains the corresponding population probabilities of these paired predictions within the stratum. The marginal rate estimates are $\widehat r_{u,ta}=(C_{10}+C_{11})/n_{ta}$ and $\widehat r_{f,ta}=(C_{01}+C_{11})/n_{ta}$. Substituting their signed gaps into Equation~\eqref{eq:target} gives $\widehat D$. Given the audit denominator, the discordant counts determine the within-group rate change because observations on which the models agree cancel in that difference. Calculating each model's absolute subgroup gap also requires its marginal rates, which the discordant counts alone do not determine. Retaining the full paired table preserves both those rates and the dependence between the models' predictions on the same observations.

An empty required audit stratum leaves insufficient sample information about a population rate that may nevertheless be defined. In this case, we return an insufficient-audit flag and the full interval allowed by the established identity windows (Appendix~\ref{app:boundary-cases}). When the conditioning event itself has zero population probability, the corresponding rate, and hence the target, is undefined. The sampling assumptions define uncertainty about the fixed histories over the prespecified horizon. Adaptive stopping, dependent observations or labels, and reuse of training observations require separate inferential justification. When model-dependent decisions alter later outcomes or their observation, causal deployment effects also require additional identification assumptions, while expected policy performance across future histories requires a separate evaluation.

\FloatBarrier
\section{Related work}
\subsection{Statistical uncertainty in fairness evaluation}
Sampling variability affects both disparity estimates and the intervals used to interpret them. \citet{lum2022debiasing} show that sampling variability can inflate disparity summaries and that correcting an estimator's bias does not necessarily calibrate its bootstrap interval. For our paired contrast, the component biases need not cancel. The joint sampling structure also matters. Under their stated assumptions for a Gaussian limit, \citet{fang2019inference} characterise bootstrap consistency through full differentiability on the limit's support. Constraints between paired estimates can make nonsmooth contributions cancel, including under prediction-function identity, so a zero component gap alone does not establish bootstrap failure.

Established alternatives address these regularity problems under different conditions. \citet{hong2018numerical} develop numerical-derivative inference using appropriately chosen step sequences, while \citet{bickel1997resampling} examine the gains and limitations of resampling fewer observations. \citet[Section~7.3 and Figure~4]{wastvedt2024intersectional} use rescaled inference for aggregates of absolute counterfactual error-rate differences and show that equal-length intervals can have different coverage. Our evaluation examines how interval construction affects both coverage and decision correctness for the paired temporal contrast. The full-size percentile--basic comparison holds observations, estimates and bootstrap draws fixed, allowing changes in audit conclusions to be traced to interval placement.

The scope and interpretation of the guarantee also matter. \citet{cherian2024auditing} distinguish certification through simultaneous subgroup bounds from flagging with false-discovery control. Our reporting distinguishes errors across all audits from errors among conclusive audits, both separate from false-discovery control over a collection of subgroup claims. \citet{barrainkua2024uncertainty} use subgroup confusion matrices for Bayesian fairness and performance comparisons, accounting for dependence across cross-validation results. Our posterior comparator retains dependence between the two models' predictions on each observation, while preserving the distinction between credible-interval interpretation and repeated-sampling coverage.

\subsection{Longitudinal audits and verdict stability}
Longitudinal evaluations establish why the effects of updating require scrutiny. \citet{bilionis2026updates} document different fairness and stability patterns under cumulative and recent-batch retraining. Monitoring during deployment additionally depends on the evolving observation process. \citet{henzinger2023runtime} derive runtime bounds for current disparity using a supplied relationship between observed interactions and changes in group expectations. \citet{chugg2023betting} provide false-alarm control across repeated inspection, with corrections for nonuniform audit sampling. We evaluate completed histories over a fixed horizon, holding fitted classifiers and realised update decisions fixed so that repeated independent audits isolate audit-sampling uncertainty.

Verdict stability concerns whether a conclusion recurs under a specified protocol. \citet{joy2026vfr} assess this through stratified bootstrap resampling, audit-size changes and hospital-grouped evaluation. Our evaluation complements such diagnostics by checking conclusions against known or numerically approximated population changes. This reference allows us to distinguish correct, incorrect and inconclusive outcomes, including cases in which a procedure produces more conclusive results together with more errors.

\subsection{Interpreting a measured disparity change}
Choices about groups, outcomes and populations define the claim before inference begins. \citet{simson2024multiverse} show that changes in group aggregation, exclusions and evaluation subsets can produce different fairness scores from unchanged predictions. \citet{pfohl2025interpretation} explain why equal subgroup performance can be misleading and why selection bias requires assumptions about the observation process. We therefore report group-specific temporal rate changes alongside the absolute-gap comparison. These findings concern a defined statistical property within an application context, consistent with the relationship between property claims and supporting evidence in the fairness assurance arguments of \citet{sabuncuoglu2025evidence}. Population specification also determines what the findings represent. \citet{wadekar2024survey} use survey and test-selection probabilities to estimate finite-population sensitivity and specificity, showing how unweighted test metrics can misrepresent those quantities. \citet{liu2026audits} examine the distribution shifts under which third-party audit conclusions remain consistent with deployment conclusions. Our descriptive Census comparison holds records, predictions and update actions fixed while varying population weights, showing how the reported disparity contrast and its group-specific interpretation depend on the population comparison being made.

\section{Sources of statistical error}
\label{sec:mechanisms}
The comparison inherits estimation errors from both models' absolute subgroup gaps. Its bias describes the average error across repeated audits, while its variance describes how much the estimates vary around their mean. Let $A_k=100T^{-1}\sum_t|g_{k,t}|$ denote the average absolute gap for each model sequence and $e_k=\widehat A_k-A_k$ its estimation error. Since the estimated contrast subtracts one component from the other, its bias and variance satisfy
\begin{align}
\operatorname{Bias}(\widehat D)&=\Ex[e_u]-\Ex[e_f],\label{eq:bias}\\
\operatorname{Var}(\widehat D)&=\operatorname{Var}(e_u)+\operatorname{Var}(e_f)-2\operatorname{Cov}(e_u,e_f).
\label{eq:variance}
\end{align}
Under the sampling model, each signed gap estimate is unbiased, meaning that its average across repeated audits equals the population gap. At a population gap of zero, positive and negative estimates cancel on average. Taking their absolute values prevents that cancellation, making nonzero estimates contribute positively. More generally, convexity gives $\Ex|\widehat g_{k,t}|\ge|g_{k,t}|$, with strict inequality when the estimated signed gap has positive probability on both sides of zero. This inflation is particularly relevant when the population gap is small relative to sampling variability, although its size also depends on the underlying rates and subgroup denominators. The difference between the two component biases can therefore displace the estimated contrast in either direction. For example, suppose the frozen model has rates of 0.5 in both groups, while the updated model has rates of 0.5 in group 0 and 0.55 in group 1. The population contrast is then an increase of $+5$ pp. With $n_1=500$ and $n_0=2000$ observations per period, finite-binomial summation gives absolute-gap biases of \fpeval{round(\ExactFrozenBias,2)} pp for the frozen model and \fpeval{round(\ExactUpdatedBias,2)} pp for the updated model. The larger bias in the frozen estimate means that too much disparity is subtracted on average. Across ten independent periods with this configuration, the contrast bias is \fpeval{round(\ExactContrastBias,2)} pp, giving an expected estimate of $+\fpeval{round(\ExactExpectedContrast,2)}$ pp despite the population increase of $+5$ pp (Appendix~\ref{app:calculations}).

Averaging across reporting periods changes the variability of this comparison without necessarily removing its bias. To examine averaging alone, consider two fixed classifiers whose joint prediction distribution is repeated across $T$ independent periods, keeping the counts within each period fixed as the total sample size grows. In this setting, the mean contrast estimate $\widehat D_T$ retains the single-period bias, while its variance is proportional to $T^{-1}$ and its standard deviation to $T^{-1/2}$. The numerical example gives
\begin{equation}
\begin{array}{c r r r}
T & \Ex[\widehat D_T]\ (\mathrm{pp}) & \operatorname{Bias}(\widehat D_T)\ (\mathrm{pp}) & \operatorname{SD}(\widehat D_T)/\operatorname{SD}(\widehat D_1)\\
1 & 3.047 & -1.953 & 1.000\\
10 & 3.047 & -1.953 & 0.316\\
\end{array}
\label{eq:temporal-illustration}
\end{equation}

At ten periods, the standard deviation is 31.6\% of its single-period value, concentrating estimates around a biased mean while the population contrast remains $+5$ pp. As intervals narrow around this biased value, they may miss the population change more often. Collecting more observations within each period can also change the bias in the absolute-gap estimates, as well as their variability across audits. Pairing affects variability through the dependence between the two models' predictions. To understand its effect within a reporting period, consider signed gaps away from zero. Taking an absolute value preserves local sampling fluctuations around a positive gap and reverses them around a negative gap, so the effect of dependence on their difference also depends on the gaps' directions. For the within-period contrast $d=|g_u|-|g_f|$, a local linear approximation gives
\begin{equation}
\operatorname{Var}(\widehat d)\approx
\operatorname{Var}(\widehat g_u)+\operatorname{Var}(\widehat g_f)
-2\operatorname{sign}(g_u)\operatorname{sign}(g_f)\operatorname{Cov}(\widehat g_u,\widehat g_f).
\label{eq:pairing}
\end{equation}
Positive signed-gap covariance reduces this approximate variance when the population gaps have the same direction and increases it when they have opposite directions. Under independent subgroup sampling, the exact signed-gap covariance is $\sum_a(q_{11,ta}-r_{u,ta}r_{f,ta})/n_{ta}$, where $q_{11,ta}$ is the probability that both models predict positively within the group/outcome stratum. Equation~\eqref{eq:pairing} requires gaps away from zero, while the general variance identity in Equation~\eqref{eq:variance} also applies at zero. The effect of pairing on precision therefore depends on the gap sizes and directions, subgroup counts and joint prediction probabilities. Appendix~\ref{app:mechanism-summary} summarises these mechanisms and their consequences.

\section{Procedures and evaluation design}
\label{sec:design}
Each audit produces an interval for the paired disparity contrast $D$ and a conclusion relative to a decision margin. Repeated audits assess interval coverage and the frequencies of correct, incorrect and inconclusive conclusions. Table~\ref{tab:review-design} summarises the study components and comparison scopes.

\begingroup
\let\designcaption\caption
\renewcommand{\caption}[1]{\designcaption{Study components and their purposes. All simulation components use ten equally weighted windows. P, B and M are defined in the text.}}
\begin{table}[htbp]\centering\small
\caption{Study components and their purposes. All simulation components use ten equally weighted windows. P/B/M are defined in Table~\ref{tab:bootstrap-methods}. The P/B comparison covers the full original evidence; reduced-size comparisons and transitions cover the specified subset.}\label{tab:review-design}
\setlength{\tabcolsep}{3pt}
\begin{tabular}{@{}L{.15\linewidth}L{.30\linewidth}L{.20\linewidth}L{.27\linewidth}@{}}\toprule
Component & Procedures and scope & Audit sizes and repeats & Purpose\\\midrule
Controlled grid & P, B, Dirichlet, projection, numerical delta; 64 configurations & $n_1=25,100,500$; $n_0=4n_1$; 2,000 audits each & Coverage and correct, incorrect and inconclusive conclusions under specified gaps and dependence\\
Original history panel & P, B, Dirichlet, projection; eight cells, twenty histories per cell, three policies, both rates & $n=1{,}250,5{,}000,20{,}000$ per window; 100 audits per unit & Reliability conditional on fitted histories; heterogeneity and identity cases\\
Reduced resampling and transitions & P, B, M; seventeen controlled configurations and two twenty-history cells, cadence policy, both rates & Same audits and sizes; four fixed margins; $\gamma=0.75$ primary & Separate interval form from resample size; identify conclusions gained, lost or made incorrectly\\
Census illustration & Point contrasts; 29 states and ten observed waves & Retained ACS records; equal record and person weights & Population specification and group outcomes\\\bottomrule
\end{tabular}\end{table}

\endgroup

\subsection{Audit conclusions and measures of reliability}
The illustrative primary margin is $\epsilon=0.5$ percentage points. For an interval $[L,U]$, the conclusion is
\begin{itemize}
\item a reduction when $U<-\epsilon$,
\item an increase when $L>\epsilon$,
\item a difference within the margin when $L\ge-\epsilon$ and $U\le\epsilon$,
\item inconclusive otherwise.
\end{itemize}
Sensitivity analyses use margins of 0.1, 0.25 and 1 percentage point. These margins classify change without determining whether remaining disparities are acceptable. Incorrect conclusions are classified as false reductions, false increases or false within-margin declarations. A false reduction declares $D<-\epsilon$ when $D\ge-\epsilon$. Its rate uses audits satisfying $D\ge-\epsilon$ as the denominator and is undefined when none are eligible. An incorrect conclusion requires an interval that excludes $D$. For this fixed comparison, valid 95\% frequentist coverage therefore bounds the error probability across all audits by 5\%. The error probability among conclusive audits can be higher. Writing $W$ for an incorrect conclusion and $C$ for any conclusive result gives $\Prob(W\mid C)=\Prob(W)/\Prob(C)$ when $\Prob(C)>0$. Both denominators are retained.

\subsection{Interval procedures}
The full-size percentile bootstrap (P) resamples each paired prediction table with replacement at its observed size. Both models' rates are calculated from the same resampled table and substituted into the contrast. The equal-tailed 95\% interval uses 10,000 draws. Full-size basic intervals (B) use these same draws, reflecting the percentile endpoints $[L_P,U_P]$ about the observed estimate to give
\[
[2\widehat D-U_P,\;2\widehat D-L_P].
\]
This preserves width while allowing different placement.  Reduced-size basic intervals (M) resample smaller paired tables and rescale the resulting errors by $\sqrt{m/N}$, where $N$ is the smallest active stratum denominator and $m=\lfloor N^\gamma\rfloor$. The primary exponent is fixed at $\gamma=0.75$. These intervals retain the full temporal horizon and use 10,000 separate draws. Appendix~\ref{app:moutn} gives the proportional allocation, integer rounding and sensitivity settings. All three bootstrap constructions use unclipped endpoints and common rules for established function identity and insufficient audits. The Dirichlet procedure instead draws paired-cell probabilities from a posterior distribution with one-half added to each observed cell count, then transforms those probabilities into $D$. Its equal-tailed 95\% credible interval uses 10,000 draws. Its posterior interpretation remains distinct from the frequentist coverage assessed through repeated audits. Projection combines Clopper--Pearson component-rate intervals through a union bound, giving coverage of at least 95\% for $D$ under the stated marginal binomial model. It permits dependence between periods provided the component counts retain that model and the identity set is fixed independently of the audit. This conservative construction can produce wide intervals because it does not fully use the paired constraints. Numerical delta uses paired bootstrap deviations to approximate the contrast's response to small perturbations of its estimated signed gaps. We evaluate an equal-tailed, unclipped basic construction with step $h_N=cN^{-1/3}$, primary $c=1$ and sensitivity constants fixed without performance-based selection. Appendix~\ref{app:methods} gives the constructions, tuning settings and boundary rules.

\subsection{Evaluation settings}
The controlled study specifies marginal rates and joint prediction probabilities to isolate bias and dependence across zero, near-zero and separated gaps, varying directions, temporal cancellation and sparse discordance. The history panel instead allows fitted functions, rates and periods with identity to vary across three updating policies. Its eight cells cross subgroup signal, model representation and drift condition. Histories are hash-selected, and calibration and evaluation histories are separate (Appendices~\ref{app:grid} and~\ref{app:histories}).

The sample sizes in Table~\ref{tab:review-design} have different meanings. Controlled counts refer to observations of the specified outcome within each group and period, with $n_0=4n_1$. History-audit sizes count all observations across groups and outcomes. Both simulation designs use ten equally weighted reporting periods. Table~\ref{tab:review-design} specifies the methods' evaluation scopes. Numerical delta is confined to the controlled study, while reduced-size basic comparisons cover selected configurations and cadence histories. The second drift setting supplies internal robustness evidence with disjoint seeds within the same simulator and calibration framework. In 181 of the 480 policy histories, the prediction functions coincide throughout, giving analytic $D=0$ targets and $[0,0]$ intervals. Summaries distinguish these identity histories, nonidentity histories and all histories. The separate descriptive Census comparison examines weighting and group outcomes using fixed predictions.

\subsection{Study summaries and numerical uncertainty}
Panel event frequencies average histories equally, with ranges describing variation between histories. Pooled error proportions among conclusive audits instead weight histories by their numbers of conclusive results. Monte Carlo uncertainty in repeated-audit frequencies is distinct from uncertainty about $D$ within an audit \citep{morris2019simulation}. Pointwise exact binomial intervals describe precision within histories or controlled configurations. Aggregate paired comparisons retain common audit eligibility, fixed history weights and covariance between procedures, yielding pointwise approximate 95\% Monte Carlo intervals (Appendix~\ref{app:aggregate}). Numerical integration approximates population references for fitted histories. Convergence checks assess sensitivity without certifying error bounds relative to the exact integrals (Appendix~\ref{app:reference}). Finite-draw endpoint approximation is assessed separately from these references and repeated-audit frequencies (Appendix~\ref{app:diagnostics}). Appendix~\ref{app:evidence} distinguishes locally specified follow-up calculations from retrospective summaries and records the chronology.

\FloatBarrier

\section{Results}
\label{sec:results}
\subsection{Interval construction and audit conclusions}
\label{sec:histories}
Across the original panel, changing from full-size percentile to basic intervals has different effects on coverage across settings. At 1,250 observations per reporting period, basic intervals have lower estimated TPR coverage in \BroadTPRLowerSmall{} of 299 nonidentity history--policy comparisons, higher coverage in nineteen and equal coverage in \BroadTPREqualSmall{}. The corresponding FPR counts are \BroadFPRLowerSmall{} lower, \BroadFPRHigherSmall{} higher and \BroadFPREqualSmall{} equal. These comparisons use common eligible audits and describe related fixed histories across the updating policies. Across the 64 controlled configurations, basic coverage is lower in 41, higher in nineteen and equal in four. These mixed directions limit any general ranking of the constructions; Appendix~\ref{app:broad-basic} reports the complete comparison across audit sizes.

To examine the consequences for audit conclusions, Table~\ref{tab:basic-main} focuses on twenty previously hash-selected subgroup-drift histories in the positive-signal setting, using the interaction representation and updating at fixed intervals. Each history has 100 independent audits with 1,250 observations per reporting period, giving 2,000 audits. This selected setting also permits comparison with reduced-size basic intervals, whose evaluation covers a narrower set of histories. Within this setting, replacing percentile with full-size basic intervals lowers estimated TPR coverage from 92.20\% to 84.55\%, with a paired Monte Carlo standard error of \AggregateSubgroupPBCoverageSE{} percentage points for the difference. At the 0.5 percentage-point margin, correct conclusive decisions increase from 78 to 376 and incorrect decisions from three to 43, including thirteen false reductions. The constructions share observations, estimates, bootstrap draws and quantiles, and have identical widths. Changing interval placement therefore produces more correct conclusions alongside more errors.

\begingroup
\setlength{\intextsep}{6pt plus 2pt minus 2pt}
\captionsetup{textformat=basicmainreadable}
\begin{table}[htbp]\centering\small
\caption{TPR audits at $n=1{,}250$: interval form and resample size. Each cell retains twenty cadence histories and 2,000 audits; the margin is 0.5 pp. Coverage is the equal history mean (\%). Correct conclusive decisions, false reduction (FR), false increase (FI), false within-margin (FW) and inconclusive (Inc.) columns are all-audit counts. Reduced-size basic uses fixed $\gamma=0.75$.}\label{tab:basic-main}
\begin{tabular}{@{}llrrrrrr@{}}\toprule
Drift & Interval & Coverage & Correct & FR & FI & FW & Inc.\\\midrule
Subgroup & Full-size percentile & 92.20 & 78 & 0 & 3 & 0 & 1,919\\
Subgroup & Full-size basic & 84.55 & 376 & 13 & 30 & 0 & 1,581\\
Subgroup & Reduced-size basic & 78.55 & 508 & 20 & 44 & 0 & 1,428\\
\midrule
Combined & Full-size percentile & 94.50 & 317 & 1 & 1 & 0 & 1,681\\
Combined & Full-size basic & 88.80 & 501 & 10 & 18 & 0 & 1,471\\
Combined & Reduced-size basic & 84.45 & 582 & 15 & 26 & 0 & 1,377\\
\bottomrule\end{tabular}
\end{table}

\endgroup

Reflecting the percentile endpoints shifts their position relative to the decision margins. Table~\ref{tab:review-transitions} follows the resulting conclusions on the same audits, showing gains in both correct and incorrect conclusions, occasional losses of correct conclusions, and further additions when resample size is reduced. All three original percentile errors persist under full-size basic, and all 43 full-size basic errors persist under reduced-size basic. 
\begin{table}[htbp]\centering\small
\caption{Which conclusions change on the same TPR audits? Each drift cell has twenty histories and 2,000 audits at $n=1{,}250$ and margin 0.5 pp. I = inconclusive, C = correct conclusive, W = incorrect conclusive. Columns show newly correct, newly incorrect and lost correct conclusions. All other transitions, including persistent errors, are in Appendix~\ref{app:transitions}.}\label{tab:review-transitions}
\begin{tabular}{@{}llrrr@{}}\toprule
Drift & Change & I$\to$C & I$\to$W & C$\to$I\\\midrule
Subgroup & P$\to$B & 299 & 40 & 1\\
Subgroup & B$\to$M & 132 & 21 & 0\\
Combined & P$\to$B & 189 & 26 & 5\\
Combined & B$\to$M & 81 & 13 & 0\\
\bottomrule\end{tabular}\end{table}

\begin{figure}[htbp]\centering
\includegraphics[width=\linewidth]{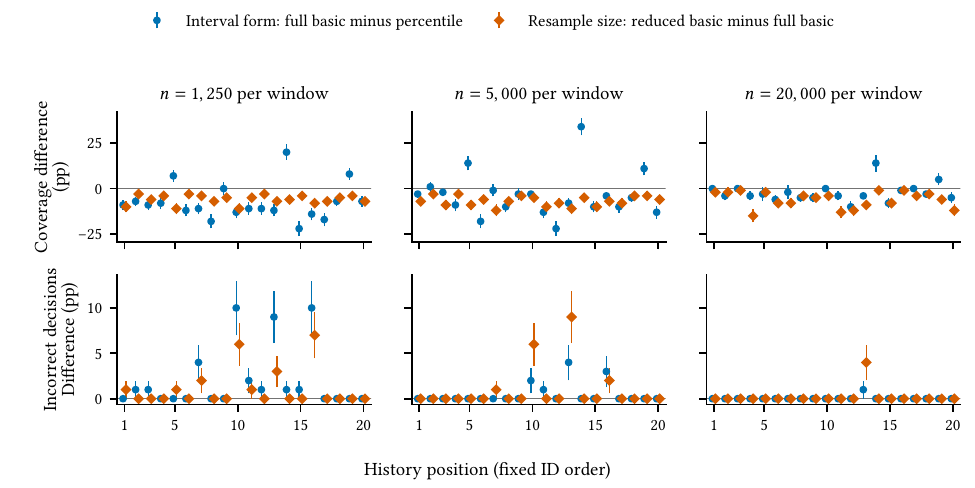}
\caption{TPR frequency differences in all twenty hash-selected subgroup-drift cadence histories. Blue circles: full-size basic minus percentile, holding data, estimates, draws and widths fixed. Orange diamonds: reduced-size minus full-size basic, retaining basic form with smaller resamples and separate draws. Negative coverage differences indicate lower coverage; positive incorrect-decision differences indicate more errors at the 0.5 pp margin. Positions retain fixed ID order, mapped in Appendix~\ref{app:history-comparator}. \textbf{Bars show Monte Carlo precision as $\pm1$ paired standard error}, estimated from 100 audits per history.}
\Description{Two rows at three audit sizes distinguish full-size basic minus percentile using blue circles from reduced-size minus full-size basic using orange diamonds. The horizontal axis has sparse labels for positions one through twenty in fixed history order. Paired Monte Carlo error bars accompany both comparisons.}
\label{fig:history-comparator}
\end{figure}

Figure~\ref{fig:disagreement} illustrates this mechanism in the original 10,000-draw calculation, where both basic intervals falsely declare reduction and percentile is inconclusive. A separate higher-draw check leaves the full-size basic verdict numerically unresolved at the decision boundary, while the reduced-size basic endpoint remains below it (Appendix~\ref{app:figure1-stability}). Across these twenty histories, reduced-size basic intervals have lower estimated TPR coverage than full-size basic intervals at all three audit sizes. Their coverage is lower than percentile coverage in \HistoryTPRLowerSmall, \HistoryTPRLowerMedium{} and \HistoryTPRLowerLarge{} histories at 1,250, 5,000 and 20,000 observations per period. Figure~\ref{fig:history-comparator} separates changes in interval form from changes in resample size, with bars showing Monte Carlo precision as one paired standard error on either side of each difference.

The additional errors affect the interpretation of conclusive results. Of the twenty reduced-size basic false reductions, eight concern a population increase beyond 0.5 percentage points and twelve concern a contrast within the margin. For percentile, full-size basic and reduced-size basic intervals, incorrect conclusions account for 3/2,000, 43/2,000 and 64/2,000 of all audits. Among conclusive audits, the fractions are 3/81 (\DecisionBootstrapWrongConclusive\%), 43/419 (10.26\%) and 64/572 (\DecisionMOutWrongConclusive\%). Pooling conclusive audits weights each history by its number of conclusive results, giving different history weights under different procedures (Appendix~\ref{app:denominators}). The paired increase from full-size to reduced-size basic intervals is \AggregateSubgroupBMErrorAllDelta{} percentage points across all audits, with an approximate 95\% Monte Carlo interval of $[\AggregateSubgroupBMErrorAllLower,\AggregateSubgroupBMErrorAllUpper]$ percentage points. Among conclusive audits, the increase is \AggregateSubgroupBMErrorConclusiveDelta{} percentage points with an interval of $[\AggregateSubgroupBMErrorConclusiveLower,\AggregateSubgroupBMErrorConclusiveUpper]$ percentage points, which includes zero. These intervals describe Monte Carlo uncertainty for the retained histories (Appendix~\ref{app:aggregate}). Effects vary with audit size and outcome rate, with smaller, mixed FPR changes in this setting.

\subsection{Internal robustness under combined drift}
The combined-drift comparison retains the positive-signal condition, representation and updating policy, using all twenty previously hash-selected histories. Evaluation and audit seeds are disjoint across drift settings, while both share the simulator and calibration procedure. This provides internal robustness evidence within the same simulation family (Appendix~\ref{app:replication}). Reduced-size basic intervals have lower TPR coverage than percentile intervals in \ReplicationTPRLowerSmall, \ReplicationTPRLowerMedium{} and \ReplicationTPRLowerLarge{} histories at the three increasing audit sizes, despite higher mean frequencies of correct conclusive decisions. Tables~\ref{tab:basic-main} and~\ref{tab:review-transitions} show that additional correct conclusions again accompany additional errors at the smallest size. For percentile, full-size basic and reduced-size basic intervals, incorrect conclusions account for 2/2,000, 28/2,000 and 41/2,000 of all audits. Among conclusive audits, the fractions are 2/319 (0.63\%), 28/529 (5.29\%) and 41/623 (6.58\%). Both denominators show an increase in the observed error fraction across these constructions.

\FloatBarrier
\subsection{Coverage and conclusions in controlled audits}
\label{sec:coverage-decisions}
\label{sec:moutn-results}
The controlled configurations show why coverage and decision correctness require separate assessment. In frozen-zero, the population contrast is $D=+5$ percentage points. At $n_1=500$ outcome-specific observations in the smaller group per period, percentile intervals cover this contrast in only 12.00\% of 2,000 audits, yet correctly declare an increase in 58.05\%, with no observed errors. An interval can lie entirely above the $+0.5$ percentage-point margin while excluding $+5$. The bias calculation in Section~\ref{sec:mechanisms} places the sampling mean of the contrast estimate at approximately $+\ExactExpectedContrast{}$ percentage points, distinct from the midpoint of any individual interval.

\begingroup
\captionsetup{textformat=directionalhighlightsreadable}
\begin{table}[t]\centering\small\setlength{\tabcolsep}{4pt}
\caption{Coverage and decisions describe different audit properties. Each row uses 2,000 saved audits and the illustrative 0.5 pp margin. Errors are partitioned by the false category declared; counts use all 2,000 audits. Zero is an observed count, not evidence of zero error probability. These configurations illustrate distinct behaviours, not the frequency of these behaviours across deployments.}\label{tab:decisions}
\begin{tabular}{@{}lrrrrrr@{}}\toprule
Procedure & Coverage & Correct & False & False & False & Inconclusive\\
 & (\%) & & reduction & increase & within & \\\midrule
\multicolumn{7}{l}{Frozen gap zero: $D=+5$ pp, $n_1=500$} \\
Full-size percentile & 12.00 & 1161 & 0 & 0 & 0 & 839 \\
Dirichlet & 11.40 & 1168 & 0 & 0 & 0 & 832 \\
Projection & 100.00 & 0 & 0 & 0 & 0 & 2000 \\
Numerical delta & 52.90 & 1897 & 0 & 0 & 0 & 103 \\
\addlinespace
\multicolumn{7}{l}{Both gaps zero: $D=+0$ pp, $n_1=25$} \\
Full-size percentile & 100.00 & 0 & 0 & 0 & 0 & 2000 \\
Dirichlet & 100.00 & 0 & 0 & 0 & 0 & 2000 \\
Projection & 100.00 & 0 & 0 & 0 & 0 & 2000 \\
Numerical delta & 88.95 & 0 & 84 & 81 & 0 & 1835 \\
\addlinespace
\multicolumn{7}{l}{Mixed gaps: $D=+2$ pp, $n_1=25$} \\
Full-size percentile & 97.55 & 0 & 3 & 0 & 0 & 1997 \\
Dirichlet & 97.75 & 0 & 2 & 0 & 0 & 1998 \\
Projection & 100.00 & 0 & 0 & 0 & 0 & 2000 \\
Numerical delta & 78.75 & 70 & 136 & 0 & 0 & 1794 \\
\addlinespace
\bottomrule\end{tabular}
\end{table}

\endgroup

In this configuration, numerical delta raises coverage to 52.90\% and the frequency of correct conclusions to 94.85\%, again with no observed errors. Full-size basic intervals yield 69.15\% coverage and 1,808 correct conclusions, retaining the percentile width of 3.55 percentage points. Reduced-size basic intervals have lower coverage of \MOutFrozenCoverage\% but more correct conclusions, at \MOutFrozenCorrect\%, and a mean width of \MOutFrozenWidth{} percentage points. Both basic constructions remain substantially below nominal coverage. Zero observed errors also leaves uncertainty about their probability. With zero errors in 2,000 independent audits, the two-sided exact 95\% Clopper--Pearson interval for the error probability across all audits has an upper endpoint of approximately 0.184\%.

Other configurations produce false reduction claims. When distinct prediction functions both have zero population gaps, $D=0$. At $n_1=25$, numerical delta has 88.95\% coverage and falsely declares reduction in \NullFalseReductionCount/2,000 audits, or \NullFalseReductionPercent\%, with a pointwise exact 95\% Monte Carlo interval of $[\NullFalseReductionLower\%,\NullFalseReductionUpper\%]$. A further 81 audits falsely declare increase, while percentile and Dirichlet intervals remain inconclusive throughout.

In mixed-gap, $D=+2$ percentage points lies beyond the increase margin, but numerical delta falsely declares reduction in \MixedFalseReductionCount/2,000 audits, or \MixedFalseReductionPercent\%, with a pointwise exact 95\% interval of $[\MixedFalseReductionLower\%,\MixedFalseReductionUpper\%]$. Changing from percentile to full-size and then reduced-size basic intervals also adds both correct and incorrect conclusions. Their false reduction counts are three, 109 and \MOutMixedFalseReduction, respectively. The same construction changes therefore have different consequences in frozen-zero and mixed-gap (Appendix~\ref{app:moutn-results}). Appendix~\ref{app:directions} reports direction-specific errors with their eligible denominators across the controlled configurations.

\subsection{Sensitivity to margins, tuning and numerical approximation}
Changing the margin can alter both the population category and the interval's conclusion while leaving coverage and width unchanged. At all four margins of 0.1, 0.25, 0.5 and 1 percentage point, moving from percentile to full-size basic and then to reduced-size basic intervals increases both correct and incorrect conclusive TPR decisions in the selected drift settings at 1,250 observations per period. Both changes also increase the observed error fraction across all audits and among conclusive audits. Relative performance is less stable across the broader panel. Excluding identity histories, the full-size basic minus percentile difference in correct-conclusion counts reverses sign relative to the primary margin in \MarginBroadReversals{} of \MarginBroadGroups{} nonempty groups defined by setting, policy, outcome rate and audit size (Appendix~\ref{app:margins}).

Fixed tuning settings also affect coverage, width and conclusions. Numerical-delta perturbations can exceed nonzero population gaps, while Dirichlet priors can materially affect results with sparse prediction disagreements. These comparisons evaluate fixed choices, with no audit-dependent tuning rule assessed (Appendix~\ref{app:tuning-interpretation}). Higher-draw checks change some interval endpoints and conclusions. They assess numerical precision conditional on retained audit observations, separately from repeated-audit coverage and decision frequencies. Appendix~\ref{app:diagnostics} reports the subset cheques, while Appendix~\ref{app:reference} describes the approximate reference to the population.

\FloatBarrier

\section{Interpreting and reporting retrospective audit claims}
\label{sec:practice}

\subsection{Conditions for interpreting an audit claim}
A decision margin should be fixed before inspecting results and reflect baseline performance, group-specific error consequences, domain expertise and affected populations' perspectives. Under a justified procedure, an interval contained in $[-\epsilon,+\epsilon]$ supports an average change within the stated margin. This concerns the average over the horizon, leaving remaining disparities and changes in individual periods for separate assessment. An inconclusive comparison remains unresolved. Interpretation depends on whether observations and labels support the sampling assumptions in Sections~\ref{sec:target} and~\ref{sec:design}. Repeated individuals, selective or delayed labels and feedback require appropriate sampling or identification assumptions. Prediction-function identity must rest on evidence independent of the audit observations. The controlled $n_1=500$ configuration requires 500 observations with the specified outcome in one group and 2,000 in the other per period. Obtaining these counts may require substantially more than 2,500 people per period in an unstratified sample, depending on outcome prevalence and group composition. Reliability assessment should therefore consider outcome-specific denominators and paired prediction disagreements alongside total sample size. The disagreements describe where the models differ on the same observations and contribute to uncertainty about their comparison.

\subsection{Population specification and group outcomes}
Population weights determine contributions to within-period rates, while temporal weights determine each period's contribution to the contrast. The descriptive Census employment example compares equal record and person weights on the same eligible, capped White-alone and Black-alone records, holding classifiers and update actions fixed. Both schemes average contrasts equally over ten observed waves within each state and then equally across the 29 states. Under updating at fixed intervals, the average FPR disparity contrast is $-0.691$ percentage points with equal record weights and $+0.528$ percentage points with person weights. The sign therefore changes with population weighting while predictions remain fixed. Mean Black-alone FPR exceeds that under the frozen classifier by 0.556 and 1.063 percentage points, respectively. The average gap reduction under equal record weights thus coexists with worse FPR for this group. These summaries describe retained records after capping and do not estimate national quantities or assess interval coverage (Appendix~\ref{app:census}). Their implications for affected people require application-specific interpretation, and disparity reduction alone does not establish broader fairness.

\subsection{Reporting the basis of an audit conclusion}
Clear reporting allows readers to scrutinise the comparison and its evidential basis. \citet[Discussion and recommendations panel, p.~11]{matos2026critical} document limited uncertainty reporting and sample-size guidance in clinical fairness evaluation. A coverage claim additionally requires an applicable validity argument or relevant validation evidence for the target and sampling design. Empirical validation should report incorrect conclusions among all audits and among conclusive audits. Separate intervals for TPR and FPR, or for several comparisons, do not by themselves establish a joint guarantee, which requires justification across the full set.

\Needspace{10\baselineskip}
An audit report should document five elements.
\begin{enumerate}
\item \textbf{Target.} Specify the estimand and population, the completed updating history and frozen comparator, the reporting periods and horizon, and the temporal and population weights.
\item \textbf{Audit data.} Describe how observations were sampled and labels obtained, giving the group--outcome denominators and full paired prediction counts. Explain how the analysis addresses dependence and any limits arising from feedback.
\item \textbf{Uncertainty.} Report the estimate, interval endpoints and nominal level, distinguishing confidence from credible intervals. Explain the construction, assumptions and justification for its interpretation, together with applicable resample sizes, tuning choices or priors, draw counts and numerical checks. State the independent evidence supporting any use of prediction-function identity.
\item \textbf{Decision.} Give the margin fixed before inspecting results, its substantive justification and the rule used to reach the reported conclusion, allowing an inconclusive outcome.
\item \textbf{Group outcomes.} Report each group's TPR and FPR under the updated and frozen classifiers and their changes, with appropriate uncertainty statements for inferential claims about those quantities.
\end{enumerate}

Table~\ref{tab:reporting} in Appendix~\ref{app:worked} illustrates how coverage properties and interval position affect the interpretation of a conclusion for one contrast.

\FloatBarrier

\FloatBarrier
\section{Conclusion and limitations}
\label{sec:limits}
More conclusive audit results did not consistently provide more reliable evidence in the evaluated settings. Following the same paired audits across interval constructions showed how additional correct conclusions could accompany additional errors, including false declarations of disparity reduction. These changes occurred even when interval widths were identical. Other controlled configurations frequently excluded the population contrast while correctly identifying an increase. Audit reliability therefore requires separate assessment of interval coverage and decision correctness for the specified comparison, sampling design and substantively justified margin.

The study assesses audit-sampling uncertainty conditional on completed updating histories, using controlled configurations and repeated independent audits of fixed logistic-regression histories. The second drift setting provides an internal robustness check within the same simulation and calibration framework. Further evaluation is needed before extending the empirical findings to other learners, updating settings or interval constructions. Expected policy performance across future updating histories and causal effects of deployment require separate evaluation.

In an applied review of a completed updating programme, the paired comparison could help assess a claim that updating reduced average disparity relative to retaining the initial model, provided the audit design and uncertainty procedure are justified for that setting. Such an assessment could contribute statistical evidence to the lifecycle risk-management and post-market monitoring activities envisaged for high-risk AI systems by the EU AI Act \citep[Articles~9(2) and~72(1)--(2)]{eu2024aiact}. Its practical value would lie in making the basis and limits of a retrospective reduction claim explicit, allowing that claim to be scrutinised alongside each group's performance and the consequences of those changes in the application.

\FloatBarrier
\label{main-text-end}
\clearpage
\section*{Ethical considerations and potential adverse impacts}
This study evaluates uncertainty about a disparity contrast between fixed prediction histories. Its simulations and descriptive analysis of existing Census data do not establish the acceptability of deploying a model or selecting an update policy. The analysis uses two-group, outcome-specific summaries, and these can obscure within-group and intersectional differences and omit harms not represented by prediction-error rates. Reducing an absolute gap can also accompany worsening group performance. Using the reporting checklist requires contextual judgement about acceptable margins and the consequences of errors, informed by the perspectives of affected populations. The checklist has yet to be evaluated through real-world deployment or stakeholder research. An inconclusive audit should not be represented as evidence that no problem exists.

\section*{Generative AI Usage Statement}
Google Gemini 3.1 Pro and Google Argon were used to assist with code checks, code execution and the development of unit tests. OpenAI GPT-6 was used to review and revise manuscript claims and wording, and to assist with implementation and testing of saved-output analyses and presentation. Human oversight included specifying and running the tests and double-checking the code. The authors retain responsibility for verifying AI-assisted contributions and for the accuracy, originality and integrity of the work.

\label{endmatter-end}
\clearpage
\bibliographystyle{ACM-Reference-Format}
\bibliography{refs,additional_refs,facct_refs}
\clearpage
\appendix
\section{Interval constructions and finite-sample calculations}
\label{app:methods}
Let $\mathcal M$ contain the $K=|\mathcal M|$ windows without established function identity; $m$ denotes the reduced resample size below. Identity windows contribute zero, but the full horizon $T$ remains in the denominator. For the paired bootstrap, independently sample
$\mathbf C^*_{ta}\sim\operatorname{Multinomial}(n_{ta},\mathbf C_{ta}/n_{ta})$
for each required stratum, calculate both marginal rates from that same table, and substitute the signed gaps into Equation~\eqref{eq:target}. The interval uses empirical 2.5th and 97.5th percentiles. The Dirichlet procedure instead draws independent $\mathbf q^*_{ta}\sim\operatorname{Dirichlet}(\mathbf C_{ta}+(a,a,a,a))$ and transforms these population probabilities directly; $a=0.5$ is the baseline. Both use 10,000 draws and linear quantile interpolation. They share the observed plug-in estimate $\widehat D$.

For numerical delta, write $\phi(g)=100T^{-1}\sum_{t\in\mathcal M}(|g_{u,t}|-|g_{f,t}|)$, $N=\min_{t\in\mathcal M,a}n_{ta}$, $r=\sqrt N$ and $h_N=cN^{-1/3}$. Paired bootstrap signed gaps give $Z^*=r(g^*-\widehat g)$ and
\begin{equation}
R^*=\frac{\phi(\widehat g+h_N Z^*)-\phi(\widehat g)}{h_N},\qquad
I_{\mathrm{ND}}=\left[\widehat D-\frac{Q_{0.975}(R^*)}{r},\ \widehat D-\frac{Q_{0.025}(R^*)}{r}\right].
\end{equation}
The construction follows numerical-derivative approaches to nonsmooth inference \citep{fang2019inference,hong2018numerical}. With proportionally growing denominators, $h_N\to0$ and $rh_N\to\infty$. These asymptotic conditions do not establish finite-sample or uniform coverage over the present configurations. The constants $c=0.5,1,2$ share paired bootstrap draws. Perturbed gaps and intervals remain unclipped. At $N=500$ and $c=1$, $h_N=0.126$ and $\sqrt N h_N=2.817$; multiplying the updated signed-gap standard deviation of approximately 0.0249 by this factor gives the perturbation scale 0.0701 discussed in Appendix~\ref{app:tuning-interpretation}.

\subsection{Paired stratified m-out-of-n construction}
\label{app:moutn}
We adapt the established reduced-resample construction \citep{bickel1997resampling}. Let $N=\min_{t\in\mathcal M,a}n_{ta}$, $m=\lfloor N^\gamma\rfloor$ and $m_{ta}=\lfloor n_{ta}m/N\rfloor$. For each active stratum independently draw
\[
\mathbf C^*_{m,ta}\sim\operatorname{Multinomial}(m_{ta},\mathbf C_{ta}/n_{ta}).
\]
Compute both marginal rates from that same table and substitute the signed gaps into the original functional with denominator $T$, obtaining $\widehat D_m^*$. No windows, groups or outcome strata are pooled. The interval is
\[
E_m^*=\sqrt{m/N}(\widehat D_m^*-\widehat D),\qquad
I_m=\left[\widehat D-Q_{0.975}(E_m^*),\ \widehat D-Q_{0.025}(E_m^*)\right].
\]
The factor follows by approximating the original root $\sqrt N(\widehat D-D)$ with the resampled root $\sqrt m(\widehat D_m^*-\widehat D)$ and dividing by $\sqrt N$. It is not a percentile interval at a smaller sample size. Under proportional growth of the active denominators, $m\to\infty$, $m/N\to0$ and $m_{ta}/m\to n_{ta}/N$, so the relative stratum information is asymptotically preserved. These rates do not establish finite-sample or uniform validity near zero gaps. The counterfactual-outcome assumptions and empirical coverage findings of \citet{wastvedt2024intersectional} do not establish validity for this paired-history target.

The reduced-size basic comparison fixes $\gamma=0.75$ for the primary analysis and $\gamma=0.65$ and $0.85$ for sensitivity analyses, without selecting among the exponents based on the results. Each calculation uses 10,000 draws, linear quantiles and no clipping, and follows the original function-identity and empty-stratum rules. Agreement between distinct functions on the observed sample does not establish identity, and resampling cannot generate prediction pairs absent from the observed table.

All 2,000 saved audits are retained for both-zero, frozen-zero, mixed-gap, local-frozen-gap and smooth same-direction configurations at each of the three smaller-group counts, plus low-rate sparse discordance at 25 and 100: seventeen configurations. The initial history comparison includes all twenty histories originally selected by hash from the positive-signal subgroup-drift cell with the interaction representation and cadence updating. It covers both rates and all three audit sizes, retaining all 100 repeats for each combination of history, rate and audit size. The cell was selected in light of earlier findings before the new comparator was calculated, and the results describe these selected settings. Across these controlled and history comparisons, the analysis produces 138,000 new intervals for the three exponents and reuses the earlier methods' endpoints unchanged.

The analysis is non-confirmatory and not externally registered. New PCG64 seeds use a separate hash namespace, leaving audit seeds unchanged. Independent stratum substreams make the first 10,000 draws identical in the 40,000-draw sensitivity check. The check uses twenty hash-selected repeats per controlled configuration and one per history/rate/size combination, for 1,380 checks across exponents. Numerical-reference exclusions and Monte Carlo variation are distinct from interval uncertainty about $D$.

An exact-enumeration unit test verifies the implementation in a smooth special case. Group 0 always predicts $(0,0)$ and group 1 predicts $(1,F)$ with $F\sim\operatorname{Bernoulli}(1/2)$, so the contrast is linear on this support: $100(1-p_f)$. The observed table for group 1 contains eight frozen positive predictions and eight negative predictions, giving an empirical resampling probability of $1/2$. With $n=16$ and $m=4$, enumerating all sixteen equally likely four-draw sequences verifies the resampled functional, zero mean of the rescaled errors, variance $100^2(1/2)(1/2)/16$, and the basic-interval endpoints.

\subsection{Projection and boundary cases}
\label{app:boundary-cases}
For $K>0$, let $[\ell_{k,ta},u_{k,ta}]$ be the Clopper--Pearson interval for a component rate, with noncoverage $\beta=\alpha/(4K)$. For a count $x$ in $n>0$ observations, the lower endpoint is zero when $x=0$ and otherwise equals $B^{-1}_{x,n-x+1}(\beta/2)$. The upper endpoint is one when $x=n$ and otherwise equals $B^{-1}_{x+1,n-x}(1-\beta/2)$. The signed-gap interval is $[\ell_{k,t1}-u_{k,t0},u_{k,t1}-\ell_{k,t0}]=[a_{k,t},b_{k,t}]$. Define
\[
L_{k,t}=\begin{cases}0,&a_{k,t}\le0\le b_{k,t},\\\min(|a_{k,t}|,|b_{k,t}|),&\text{otherwise},\end{cases}
\qquad U_{k,t}=\max(|a_{k,t}|,|b_{k,t}|).
\]
The projected disparity interval is
\[
\frac{100}{T}\left[\sum_{t\in\mathcal M}(L_{u,t}-U_{f,t}),\quad
\sum_{t\in\mathcal M}(U_{u,t}-L_{f,t})\right].
\]
The union bound gives joint inclusion of all component rates with probability at least $1-\alpha$ under the conditional binomial sampling model; joint inclusion implies inclusion of $D$. Dependence between paired rates does not invalidate that bound. The component box may contain combinations excluded by a paired joint region and is consequently conservative.

When all windows have established identity, every method returns $[0,0]$ analytically. Sample agreement between distinct functions retains a degeneracy flag rather than invoking this identity rule. A missing required audit denominator gives the full range permitted by established identity, $[-100K/T,100K/T]$ pp, and an insufficient-audit flag. This reduces to $[-100,100]$ pp when $K=T$; identity windows contribute zero. Undefined population conditioning events are distinct from empty sample strata. Bootstrap and posterior component intervals describe marginal group-specific changes, without implying simultaneous 95\% coverage of every displayed quantity.

\subsection{Finite-binomial absolute-gap bias}
\label{app:calculations}
For independent $B_a\sim\operatorname{Binomial}(n_a,p_a)$, the absolute-gap bias in proportions is
\begin{equation}
\sum_{i=0}^{n_1}\sum_{j=0}^{n_0}\left|\frac{i}{n_1}-\frac{j}{n_0}\right|
\binom{n_1}{i}p_1^i(1-p_1)^{n_1-i}
\binom{n_0}{j}p_0^j(1-p_0)^{n_0-j}-|p_1-p_0|.
\end{equation}
The finite sum is evaluated in double precision, without simulation or an asymptotic approximation. Within-observation covariance between updated and frozen predictions affects the variance of their difference, but neither model's marginal bias. Table~\ref{tab:exact_bias} reports the finite-binomial calculations.
\begingroup
\captionsetup{textformat=finalcaptionexactbias}
\begin{table}[!htbp]\centering\small
\caption{Finite-binomial expectation for the frozen-zero case, in pp. The updating and frozen group-0 rates are 0.5; their group-1 rates are 0.55 and 0.5. These deterministic calculations check existing configurations and add no experimental cases.}\label{tab:exact_bias}
\begin{tabular}{@{}rrrrr@{}}\toprule
$n_1$ & $n_0$ & Frozen bias & Updating bias & Contrast bias\\\midrule
25 & 100 & 8.934213 & 4.791676 & -4.142537 \\
100 & 400 & 4.461989 & 1.128592 & -3.333398 \\
500 & 2000 & 1.994861 & 0.041854 & -1.953008 \\
\bottomrule\end{tabular}
\end{table}

\endgroup

\subsection{Mechanism summary}
\label{app:mechanism-summary}
\begin{table}[htbp]\centering\small
\caption{Statistical mechanisms and their consequences for the paired temporal audit.}\label{tab:contribution}
\begin{tabular}{@{}L{0.29\linewidth}L{0.65\linewidth}@{}}\toprule
Established principle & Evidence for interpreting the audit\\\midrule
Convexity biases absolute estimates upward & The difference between component biases can be large or nearly cancel. In the preselected history, appreciable component errors coexist with small contrast bias (Appendix~\ref{app:geometry}).\\
Covariance changes contrast variance & With marginal rates fixed, the empirical variance ratios for prediction covariance 0.12 versus zero are \PairingSame{} for same-direction gaps and \PairingOpposite{} for opposite directions, estimated from 2,000 audits per configuration at $n_1=500$ and $n_0=2000$.\\
Averaging reduces independent noise & Repeated gap configurations retain common bias across windows. Undercoverage can accompany correct conclusive decisions about a change exceeding the margin (Section~\ref{sec:results}).\\
Valid coverage bounds wrong interval-based decisions & The converse does not hold. The experiments separate magnitude errors, false reductions and unresolved comparisons, including failures of a specified numerical-delta construction.\\\bottomrule
\end{tabular}
\end{table}

\section{Controlled design, uncertainty and sensitivity}
\label{app:grid}
\begin{figure}[htbp]
\centering
\includegraphics[width=\linewidth]{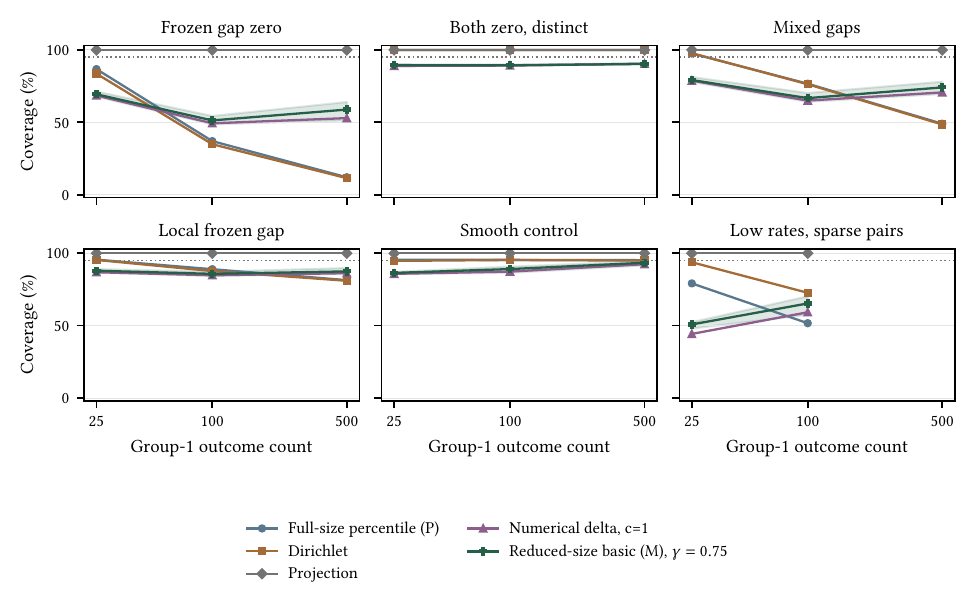}
\caption{Coverage across six controlled configuration families. Curves show percentile, Dirichlet, projection, numerical-delta and primary reduced-size basic intervals. Shading spans the three fixed exponents, not a confidence band. Each configuration uses the same 2,000 audits, $n_0=4n_1$ and ten windows; sparse-discordance cases use 25 and 100 only. The dotted reference is 95\%.}
\Description{Six panels compare coverage of the original four procedures with the paired m-out-of-n bootstrap. A shaded envelope spans its fixed exponent choices. The additional construction improves coverage in some configurations without restoring nominal coverage across them all.}
\label{fig:coverage}
\end{figure}

The original 42 configurations arise from the twelve gap rows in Table~\ref{tab:grid}, each at $n_1=25,100,500$, plus covariance-zero versions of the same-direction and opposite-direction rows at each size. Every configuration has $n_0=4n_1$, $T=10$ and 2,000 independent audits. Group-0 rates are 0.5 for both models; group-1 rates add the specified signed gap. For distinct functions, within each group and window,
\[
q_{11}=p_up_f+c,\quad q_{10}=p_u-q_{11},\quad q_{01}=p_f-q_{11},\quad
q_{00}=1-p_u-p_f+q_{11},
\]
where $c=0.12$ unless zero covariance is specified. Known identity instead has $q_{11}=p_u=p_f$ and zero discordance. The four cells are strictly positive in the nonidentity covariance-0.12 rows; their prediction correlations range from 0.480 to approximately 0.535. Targets, including exact margin cases, are stored as rational values.

\begin{table}[htbp]\centering
\caption{Original controlled cases. Gaps are proportions and repeat across windows unless stated otherwise. Each row is evaluated at three denominator sizes.}\label{tab:grid}
\begin{tabular}{@{}lrr@{}}\toprule
Case & Frozen gap & Updating gap\\\midrule
Same direction, away from zero & 0.150 & 0.100\\
Opposite directions & 0.150 & $-0.100$\\
Frozen gap zero & 0.000 & 0.050\\
Updating gap zero & 0.050 & 0.000\\
Both zero, distinct functions & 0.000 & 0.000\\
Frozen gap near zero & 0.005 & 0.055\\
Updating gap near zero & 0.055 & 0.005\\
Reduction margin & 0.150 & 0.145\\
Increase margin & 0.150 & 0.155\\
Equal nonzero, distinct functions & 0.150 & 0.150\\
Temporal cancellation & 0.150 & 0.170 then 0.130\\
Known identical functions & 0.150 & 0.150\\\bottomrule
\end{tabular}
\end{table}

The extension adds 22 configurations. Mixed gaps use four windows with $(g_f,g_u)=(0,0.05)$, three with $(0.005,0.055)$ and three with $(0.15,0.10)$, giving $D=2$ pp. The local sequence uses $g_f=\sqrt{0.25/n_1+0.25/n_0}$ and $g_u=g_f+0.05$, giving $D=5$ pp; the evaluated square root is stored as a decimal rational. Both types use group-0 rates 0.5 and covariance 0.12 at all three sizes.

The sixteen configurations used to assess prior sensitivity have group-0 and frozen group-1 rates $b\in\{0.02,0.5,0.98\}$. Updating group-1 rates are $b+0.05$ at low and central rates and $b-0.05$ at high rates. Set $q_{11}=p_up_f+\eta\{\min(p_u,p_f)-p_up_f\}$, with $\eta=0.5$ or $0.95$. The three baseline rates, two values of $\eta$ and two sample sizes ($n_1=25,100$) yield twelve nonzero-target configurations. Four further configurations use low or high rates, zero gaps and $\eta=0.95$ at these two sizes. All paired-cell probabilities are strictly positive, and the prediction functions are distinct. Complementing the predictions maps the low-rate design to the high-rate design, while their audit streams are independent.

Prior sensitivity covers 31 configurations: the fifteen original frozen-zero, updating-zero, both-zero, opposite-direction and known-identity settings, plus all sixteen new prior stresses. Under $\operatorname{Dirichlet}(a,a,a,a)$, each marginal rate and total discordance has a $\operatorname{Beta}(2a,2a)$ prior. A marginal success count $x$ has posterior mean $(x+2a)/(n+4a)$; discordance has mean $(C_{01}+C_{10}+2a)/(n+4a)$. At $n=25$, the marginal prior-mean weights for $a=0.1,0.5,1$ are 1.6\%, 7.4\% and 13.8\%. In the low-rate null with $b=0.02$, $\eta=0.95$, discordance probability is $2(1-\eta)b(1-b)=0.00196$, explaining the expected count 0.049 in a smaller-group/window stratum.

\subsection{Monte Carlo uncertainty and diagnostic assessments}
\label{app:diagnostics}
The numerical assessments distinguish sampling variation across fresh audits of a fixed history, finite-draw endpoint approximation for fixed audit counts, and numerical approximation of population reference rates. Appendix~\ref{app:figure1-stability} reports the targeted precision assessment of Figure~\ref{fig:disagreement}.

In the frozen-zero configuration at $n_1=500$, primary numerical-delta coverage is 42.10 percentage points below the nominal 95\% level. Its gain over full-size percentile intervals is 40.90 percentage points (paired Monte Carlo standard error 1.11 pp).

For $k$ events in $R>0$ eligible independent audits, pointwise exact 95\% Monte Carlo intervals use Clopper--Pearson beta quantiles with tail probability 0.025. These intervals describe uncertainty in event probabilities, not uncertainty about $D$. Zero events have a positive upper bound; $R$ events have a lower bound below one. No eligible audits means an undefined rate. Across histories, the report first calculates each conditional frequency and then averages histories equally; ranges describe heterogeneity rather than uncertainty in that mean.

The original diagnostic assessment uses $42\times3\times2=252$ one-sided bounds, and the extension uses $64\times4\times2=512$. A method--configuration combination qualifies only if its upper bound on noncoverage is at most 7\% and its lower bound on the frequency of correct conclusive decisions is at least 80\%. The tail probabilities are $0.05/252$ for the original family and $0.05/512$ for the extension. Failure to qualify does not by itself establish undercoverage, and qualifying fractions describe this designed grid. Alternative step sizes, prior concentrations and margins are assessed as sensitivity analyses, with no preferred setting selected from their results.

The extension contains 252,000 reused baseline intervals and 640,000 new intervals, totalling 892,000. Of the new intervals, 596,000 are stochastic and 44,000 are deterministic projection intervals. A fixed 1\% subset of the stochastic intervals (5,960) was recalculated with 40,000 draws. Twenty primary-margin decisions changed, with eight becoming inconclusive and twelve becoming conclusive and no switches between conclusive categories. Three changes introduced an incorrect conclusion and two removed one. The maximum endpoint shift was 0.43 pp, and all 256 method/configuration diagnostic classifications were unchanged. Table~\ref{tab:numerical-transitions} reports the changes by method.

For the highlighted mixed-gap numerical-delta result, substituting the twenty checked intervals into the 2,000 primary outputs changes the number of false reductions from 136 to 135. The corresponding substitution for reduced-size basic intervals leaves the count at 129. Each check covers twenty intervals, two of which declare a false reduction, while the other 1,980 intervals retain their original draw count. The projection intervals in the worked audit and Table~\ref{tab:reporting} use deterministic component quantiles and are independent of the resampling draw count.

The $10^{-5}$ pp tolerance flags comparisons in which the population reference lies near an interval endpoint or decision boundary. It does not control error from using finitely many resampling draws. Across the original full panel, five method/history/size coverage records each exclude one numerically unresolved repeat, comprising two full-size percentile and three posterior records. The population references for all primary decision classifications are numerically resolved. Applied audits require a prespecified endpoint tolerance and escalation rule, so a result should remain inconclusive when numerical precision is insufficient to classify the interval. Any stopping rule also requires separate justification.

\begingroup
\captionsetup{textformat=numericaltransitions}
\renewcommand{\reporttable}[1]{\input{presentation/#1}}
\begin{longtable}{@{}llrrr@{}}
\caption{Saved primary-margin decision changes by method. C denotes a conclusive category and I an inconclusive result. Columns count C$\to$I, I$\to$C and changes between conclusive categories. Unchanged methods are retained in the machine-readable record.}\label{tab:numerical-transitions}\\
Study & Method & C$\to$I & I$\to$C & C$\to$C\\\toprule\endfirsthead
Study & Method & C$\to$I & I$\to$C & C$\to$C\\\toprule\endhead
Comparator & Reduced-size basic ($\gamma=0.75$) & 1 & 0 & 0\\
Extension & Full-size percentile & 0 & 2 & 0\\
Extension & Dirichlet ($a=0.5$) & 1 & 0 & 0\\
Extension & Dirichlet ($a=0.1$) & 2 & 0 & 0\\
Extension & Dirichlet ($a=1$) & 1 & 1 & 0\\
Extension & Numerical delta ($c=0.5$) & 3 & 3 & 0\\
Extension & Numerical delta ($c=1$) & 1 & 4 & 0\\
Extension & Numerical delta ($c=2$) & 0 & 2 & 0\\
\bottomrule\end{longtable}

\endgroup

For all 64 controlled configurations and the four original procedures, we calculate coverage with exact 95\% Monte Carlo intervals, mean width in pp, percentages of correct conclusive decisions and inconclusive audits, and counts of incorrect conclusive decisions out of 2,000 audits at the 0.5 pp margin. The C and R diagnostic labels refer to the coverage and correct-decision criteria in the separate 512-bound family described above. Figure~\ref{fig:coverage} provides the controlled coverage overview. Identity rows have analytically known $D=0$, independently of the empirical binomial bounds.

\FloatBarrier
\subsection{Interpreting tuning sensitivity}
\label{app:tuning-interpretation}
The sensitivity analyses show how fixed numerical-delta step sizes, Dirichlet prior concentrations and reduced-size exponents affect coverage, interval width and audit conclusions.

The numerical-delta alternative improves coverage in several configurations without providing uniformly reliable coverage across the grid (Figure~\ref{fig:coverage}). In the frozen-zero case at $n_1=500$, coverage is 65.55\%, 52.90\% and 40.05\% for $c=0.5,1,2$; mean widths are 3.30, 2.77 and 2.48 pp. For $c=1$, the perturbation's approximate standard-deviation scale is 0.0701, exceeding the nonzero updated gap of 0.05. The asymptotic step-size conditions alone do not resolve this finite-sample sensitivity.

Prior strength changes coverage and the frequency of correct conclusive decisions. For low rates, sparse discordance and $n_1=25$, Dirichlet coverage is 84.05\%, 93.80\% and 97.95\% for $a=0.1,0.5,1$, while the frequency of correct conclusive decisions falls from 61.70\% to 33.20\%. At $n_1=100$, coverage remains 57.05\%, 72.70\% and 84.50\% despite correct increase decisions above 99\%. The prior discordance pseudocount can exceed the expected count of discordant observations. Changing the concentration alters both shrinkage and dependence, so these results do not separate their effects (Appendix~\ref{app:grid}).

Step-size sensitivity evaluates all 64 numerical-delta configurations at $c=0.5,1,2$, and prior sensitivity evaluates all 31 Dirichlet configurations at $a=0.1,0.5,1$. We calculate coverage, mean width in pp, percentages of correct conclusive decisions and counts of incorrect conclusive decisions out of 2,000 audits at the primary 0.5 pp margin, together with inconclusive frequencies, pointwise Monte Carlo intervals and paired standard errors.

\FloatBarrier

\section{Direction-specific decision errors}
\label{app:directions}
We calculate directional errors from saved interval endpoints using the decision rule in Section~\ref{sec:design}. Exact rational targets determine true categories, including targets exactly on a margin. The analysis verifies every saved decision against its endpoints and reconciles coverage, correct, wrong and inconclusive counts with the frozen summaries for every evaluated method and margin. Counts of false reductions, false increases and false within-margin declarations partition all incorrect conclusive decisions. Each false-declaration rate uses only audits in which that declaration would be false and is undefined when the eligible denominator is zero. The directional comparisons cover all four evaluated margins, with eligible denominators and pointwise exact binomial bounds calculated separately for each controlled configuration.

\section{Retained-history specification}
\label{app:histories}
The eight cells cross two subgroup-signal conditions, two representations and two drift regimes. Let $A\sim\operatorname{Bernoulli}(0.2)$, independently of $Z\sim N_{10}(0,\Sigma)$, and let $I_c$ indicate combined drift. With $d_t=0.05\max(t,0)$,
\begin{align}
X&=Z+Am+I_c d_t(e_0+e_1-Ae_2-Ae_3),\\
\Pr(Y=1\mid X,A)&=\sigma\!\left[\alpha_c+X^\top\{(1+\gamma_c A-d_t A)\beta+I_c d_t e_4\}\right],
\end{align}
where $\sigma$ is the logistic function, $e_j$ a coordinate unit vector, and
\[
m=(0.35,-0.20,0.15,0.10,0,0,0,0,0,0)^\top,
\quad\beta=(0.55,-0.40,0.35,0.25,-0.20,0.18,-0.12,0.08,0.05,-0.04)^\top.
\]
The two independent five-coordinate covariance blocks have unit diagonal and off-diagonal correlation 0.3. Reversed signal uses $\gamma_c=-1.375$ and historical regime-specific intercepts approximately $-0.190323$ and $-0.196349$. Positive signal uses $\gamma_c=-0.25$ and a baseline intercept calibrated to marginal prevalence 0.45, without targeting parity or a policy contrast. Changing signal changes baseline gaps, so factor comparisons do not isolate causal effects of signal direction or representation.

The learner uses either $X$ alone or $[X,A,AX]$, with L2-regularised logistic regression, $C=1$, an unpenalised intercept, no sample or class weights, L-BFGS, tolerance $10^{-4}$, maximum 500 iterations, random state zero and a 0.5 cutoff. Simulation predictors are not standardised. The initial fit uses 5,000 observations. Refits at boundary $t$ use deployment windows $\max(0,t-3),\ldots,t-1$, excluding the initial training sample. A boundary-2 fit has 10,000 observations and subsequent full fits 15,000. One-window label delay makes labels from window $t-1$ available at boundary $t$.

Cadence refits at boundaries 3, 6 and 9. The loss policy uses upward cumulative log-loss departures from its window-0 reference; the gap-change policy monitors departures in both directions for signed TPR and FPR gaps. Boundaries 2--9 are eligible. A strict threshold crossing triggers a refit and resets all carried accumulators for that policy; the reference does not reset. Simulation scales are one and allowances zero. Thresholds use 300 separate stationary calibration histories to target a 10\% chance of any action within ten windows, without matching refit budgets. The historical evaluation contains 400 histories per cell and stationary counterparts; the audit retains the original hash-selected twenty per cell.

Audit sampling is independent of the observations used to fit models or trigger updates. The three total sample sizes are nested within each repeat, so comparisons across sizes are paired rather than independent studies. Each fixed history has 100 independent repeats per size. The 480 policy histories include 181 with established identity throughout. We report summaries separately for nonidentity histories, histories with identity throughout, and all histories. Summaries for empty sets of nonidentity histories remain undefined. Population rates use the frozen numerical calculations with $10^{-5}$ pp comparison tolerance. The history-generating study is non-confirmatory.

\subsection{Population references and numerical accuracy}
\label{app:reference}
The reference rates are population expectations conditional on the fitted classifiers, rather than estimates from a reference Monte Carlo sample. For group $a$ and window $t$, let $X\sim N(\mu_{ta},\Sigma)$, $p_{ta}(x)=\sigma(\alpha_{ta}+\beta_{ta}^{\top}x)$ be the outcome probability from the stated simulator, and let $h_{k,ta}(x)=1\{b_{k,ta}+v_{k,ta}^{\top}x\ge0\}$ be the fitted classifier. Then
\[
r_{k,ta}^{\mathrm{TPR}}=\frac{\mathbb E[h_{k,ta}(X)p_{ta}(X)]}{\mathbb E[p_{ta}(X)]},\qquad
r_{k,ta}^{\mathrm{FPR}}=\frac{\mathbb E[h_{k,ta}(X)\{1-p_{ta}(X)\}]}{\mathbb E[1-p_{ta}(X)]}.
\]
The group-specific effective affine coefficients include the interaction representation. Thus these expectations use the actual fixed prediction functions, the common cutoff and the window-specific evaluation population.

The original integrator conditions the Gaussian classifier score on the Gaussian outcome logit. Writing that logit as $m_y+s_y Z$, $Z\sim N(0,1)$, the conditional probability of a positive prediction is $H(Z)=\Phi\{(m_f+\lambda Z)/\tau\}$, where $\lambda=\operatorname{Cov}(v^\top X,\beta^\top X)/s_y$ and $\tau^2=\operatorname{Var}(v^\top X)-\lambda^2$. Numerators and denominators therefore become one-dimensional normal integrals of $\sigma(\pm(m_y+s_y z))H(z)$ and $\sigma(\pm(m_y+s_y z))$, respectively. Zero-variance scores use their analytic limits.

SciPy's adaptive QUADPACK \texttt{quad} evaluates these integrals on $[-12,12]$, with both absolute and relative tolerances $10^{-8}$ for a coarse calculation and $10^{-11}$ for the fine reference, and a subdivision limit of 200. Additional breakpoints surround the classifier boundary at offsets $0,\pm1,\pm4,\pm12$ times its conditional transition width. Integration warnings are treated as errors. Unresolved outcome denominators and invalid variances are rejected, allowing only negative roundoff within a relative $10^{-12}$ variance tolerance.

The historical numerical checks include 1,010,720 component-rate evaluations, covering calibration and candidate models. The largest absolute coarse--fine rate change is $3.504\times10^{-11}$ as a proportion. By the reverse triangle inequality, component bounds $\delta_{k,ta}$ imply
\[
|D-\widetilde D|\le\frac{100}{T}\sum_{t\in\mathcal M}\sum_{k,a}\delta_{k,ta}
\le400\frac{K}{T}\delta
\]
if every active component error is at most $\delta$. Applying this inequality to the observed coarse--fine changes bounds the difference between those two numerical calculations by $1.402\times10^{-8}$ pp for any comparison covered by these checks. This is a convergence comparison, not a certified bound on error relative to the exact population integral. The largest saved propagated contrast sensitivity is $8.160\times10^{-9}$ pp; it combines coarse--fine changes and propagated QUADPACK error estimates. All eighty history--rate targets in the forty-history comparison match the historical fine references within $10^{-10}$ pp. Table~\ref{tab:reference-checks} summarises the numerical checks.

\begingroup
\captionsetup{textformat=referencechecks}
\begin{table}[htbp]\centering\small
\caption{Recovered historical numerical-reference checks. The rate ledger covers the parent numerical evaluation, including calibration and candidate models; it is broader than the selected audit histories. Sensitivities are estimated, not certified bounds. No integration was rerun for this revision.}\label{tab:reference-checks}
\begin{tabular}{@{}lr@{}}\toprule Check & Value\\\midrule
Component-rate evaluations & 1,010,720\\
Comparator history--rate targets reconciled & 80\\
Largest coarse--fine rate change (proportion) & $3.504\times10^{-11}$\\
Implied coarse--fine contrast-change bound (pp) & $1.402\times10^{-8}$\\
Largest saved propagated sensitivity (pp) & $8.160\times10^{-9}$\\
Original reference-comparison tolerance (pp) & $10^{-5}$\\\bottomrule
\end{tabular}\end{table}

\endgroup

The $10^{-5}$ pp comparison tolerance exceeds the observed numerical sensitivities. It classifies near-boundary comparisons as unresolved and is not a certified integration-error bound. Coverage is unresolved when a nonidentity audit interval endpoint lies within this tolerance of the reference, while category correctness is unresolved when the reference lies within the tolerance of a decision margin. An unresolved audit is omitted only from the affected denominator. The original full panel excludes one repeat from each of five method--history--size coverage records, while all primary decision references are resolved. The additional comparison of full-size percentile, full-size basic and reduced-size basic intervals has one percentile coverage exclusion (Appendix~\ref{app:transitions}) and no decision exclusions at any of its four margins.

For a future reference with a justified bracket $[D^-,D^+]$, coverage is definite if that bracket lies inside the audit interval, noncoverage is definite if the two are disjoint, and other cases are numerically unresolved. The historical convergence checks do not provide a certified bracket of this kind.

\section{Initial reduced-size basic results}
\label{app:moutn-results}
This appendix reports the initial controlled and subgroup-drift comparisons; Appendix~\ref{app:replication} reports the combined-drift extension.
\begin{table}[htbp]\centering\small
\caption{Paired stratified reduced-size basic intervals on the same 2,000 controlled audits as Table~\ref{tab:decisions}. The primary exponent is $\gamma=0.75$; both sensitivity exponents are reported. Decisions use the 0.5 pp margin. FR/FI/FW are false reduction/increase/within-margin counts; correct conclusive decisions and inconclusive audits are counts. Complete results include additional controls and exact Monte Carlo intervals.}\label{tab:moutn-main}
\begin{tabular}{@{}lrrrrrr@{}}\toprule
Case ($n_1$) & $\gamma$ & Coverage (\%) & FR/FI/FW & Correct & Inconclusive & Width (pp)\\\midrule
Frozen zero (500) & 0.65 & 51.30 & 0/0/0 & 1897 & 103 & 2.73\\
Frozen zero (500) & 0.75 & 58.80 & 0/0/0 & 1887 & 113 & 2.95\\
Frozen zero (500) & 0.85 & 64.00 & 0/0/0 & 1870 & 130 & 3.19\\
Both zero (25) & 0.65 & 88.70 & 87/86/0 & 0 & 1827 & 11.31\\
Both zero (25) & 0.75 & 89.50 & 85/82/0 & 0 & 1833 & 11.66\\
Both zero (25) & 0.85 & 90.05 & 75/73/0 & 0 & 1852 & 12.06\\
Mixed gaps (25) & 0.65 & 77.90 & 143/0/0 & 79 & 1778 & 11.37\\
Mixed gaps (25) & 0.75 & 79.20 & 129/0/0 & 67 & 1804 & 11.74\\
Mixed gaps (25) & 0.85 & 81.25 & 120/0/0 & 61 & 1819 & 12.15\\
\bottomrule\end{tabular}
\end{table}

Table~\ref{tab:conditional-coverage} reports the subgroup-drift cadence example. The fixed-exponent comparison evaluates $\gamma=0.65,0.75,0.85$ in the seventeen controlled configurations and the twenty retained subgroup-drift cadence histories, with both rates and all three audit sizes in that cell. The primary exponent remains $\gamma=0.75$ and the primary margin remains 0.5 pp. The comparisons cover all four margins and every original comparator, with event-specific eligible denominators, pointwise Monte Carlo intervals, degenerate-resampling counts and controlled paired differences. No simultaneous diagnostic-family threshold is applied to reduced-size basic intervals.

Holding the audit samples fixed, increasing the draw count from 10,000 to 40,000 changes \MOutNumericalChanges{} of the 1,380 primary-margin decisions, with a maximum endpoint shift of \MOutMaximumEndpointShift{} pp. The 40,000-draw streams extend the original 10,000-draw streams exactly.

\begingroup
\captionsetup{textformat=finalcaptionpanelconditionalsummary}
\begin{table}[!htbp]\centering\small
\caption{Conditional coverage and correct conclusive decisions for cadence in the positive-signal interaction cell under subgroup drift. All sizes use the same 20 acting histories and 100 audits per history. Posterior coverage concerns TPR; percentile decisions use the 0.5 pp margin. Percentages round exact count fractions half up. History 259 is highlighted retrospectively and has the lowest posterior TPR coverage at all three sizes. Its brackets are pointwise exact 95\% binomial Monte Carlo intervals, not selection-adjusted or population-of-histories intervals.}\label{tab:conditional-coverage}
\setlength{\tabcolsep}{4pt}
\begin{tabular}{@{}rrrrr@{}}\toprule
Audit size & \multicolumn{2}{c}{Posterior TPR coverage} & \multicolumn{2}{c}{Percentile correct (\%)}\\
per window & Mean (\%) & History 259: count [\%] & TPR & FPR\\\midrule
1,250 & 91.9 & 58/100 [47.7, 67.8] & 3.9 & 67.2\\
5,000 & 91.3 & 60/100 [49.7, 69.7] & 29.8 & 76.4\\
20,000 & 93.3 & 74/100 [64.3, 82.3] & 56.6 & 85.5\\
\bottomrule\end{tabular}
\end{table}

\endgroup
\FloatBarrier

\section{Retrospective associations across retained histories}
\label{app:geometry}
The broader panel contains 299 nonidentity policy histories, each assessed for TPR and FPR, giving 598 history--rate comparisons. At $n=5{,}000$, mean percentile coverage is 93.67\%, 95.57\% and 94.99\% in retrospective smallest-gap bands of at most 1 pp, over 1 to 5 pp and over 5 pp, with incorrect-conclusion frequencies of 0.12\%, 0.15\% and 0.23\%, respectively. These retrospective gap bands and rank associations describe the selected panel. Policies and rates share histories and audit inputs, while bias cancellation, covariance and active-window counts vary together. Low pooled error frequencies can coexist with poor coverage in individual histories.

Each of the 598 nonidentity comparisons contains 100 audits per size for each of the three baseline procedures. Five coverage records use 99 eligible audits after excluding one numerically unresolved repeat under the original tolerance rule, while every decision record uses all 100 audits. The 181 identity histories remain separate.

\FloatBarrier

\section{The worked audit}
\label{app:worked}
History 144 belongs to the positive-signal, $X$-only, combined-drift cell under the gap-change policy. Selection as the first hash-ranked history preceded interval calculation. Repeat 0 at 5,000 observations per window yields $\widehat D=0.050$ pp. The five windows without established function identity contribute to the numerator, while all ten windows remain in the denominator. The average of the within-window absolute-gap differences gives the estimate before display rounding.

\begin{figure}[htbp]
\centering
\fbox{\begin{minipage}{0.93\linewidth}
\textbf{What can the auditor conclude?}\par\smallskip
Two procedures applied to the same paired TPR audit; 5,000 observations per window.\par
\begin{center}
\begin{tabular}{@{}lr@{}}
Estimated updating-minus-frozen disparity & $+0.050$ pp\\
Full-size percentile interval (nominal 95\%) & $[-0.627,\;0.869]$ pp\\
Projection interval (at least 95\% coverage) & $[-7.645,\;7.698]$ pp\\
Illustrative decision margins & $-0.5$ and $+0.5$ pp\\
\end{tabular}
\end{center}
Both intervals cross the reduction and increase thresholds, so both audits are inconclusive. The near-zero estimate alone does not establish a reduction or a change within the margin.

\end{minipage}}
\caption{History 144, audit repeat 0, selected by hash ordering before interval calculation. All changes are in percentage points (pp). The 5,000 observations are the total per window and inference uses only positive-outcome observations in each group. Table~\ref{tab:worked-intervals} gives all three baseline intervals.}
\Description{Boxed example showing an estimated disparity change of plus 0.050 percentage points, a bootstrap interval from minus 0.627 to plus 0.869, and a projection interval from minus 7.645 to plus 7.698. Both procedures return an inconclusive result at the plus or minus 0.5 margin.}
\label{fig:worked}
\end{figure}

\begin{table}[htbp]\centering\small
\caption{Worked-audit intervals for the TPR comparison, in pp. Each disparity interval crosses the $-0.5$ and $+0.5$ pp decision boundaries. The group-specific intervals describe average changes in TPR. Full-size percentile and posterior component intervals are marginal and do not establish joint coverage of all reported quantities.}\label{tab:worked-intervals}
\begin{tabular}{@{}llrrr@{}}\toprule
Procedure & Interval & Disparity & Group 0 change & Group 1 change\\\midrule
Percentile & Confidence & $[-0.627,0.869]$ & $[-0.244,0.401]$ & $[-0.186,1.048]$\\
Dirichlet & Credible & $[-0.613,0.896]$ & $[-0.250,0.390]$ & $[-0.174,1.052]$\\
Projection & Confidence & $[-7.645,7.698]$ & $[-3.513,3.655]$ & $[-6.701,7.547]$\\\bottomrule
\end{tabular}
\end{table}
\begin{table}[htbp]\centering\small
\caption{Interpreting an audit at the illustrative $\pm0.5$ pp margins. The conclusive nonidentity example was selected retrospectively to illustrate the decision rule. All interpretations assume the stated sampling model.}\label{tab:reporting}
\begin{tabular}{@{}L{0.25\linewidth}L{0.31\linewidth}L{0.37\linewidth}@{}}\toprule
Coverage and decision & Audit example & Interpretation\\\midrule
Coverage guarantee; conclusive & History 2, FPR, repeat 0, $n=20{,}000$: projection gives $[\ResolvedProjectionLower,\ResolvedProjectionUpper]$ pp. & This supports a reduction of more than 0.5 pp in the average absolute FPR gap.\\
Coverage guarantee; inconclusive & Projection in the worked audit gives $[-7.645,7.698]$ pp. & The interval spans both decision boundaries, so the comparison remains inconclusive.\\
Empirical undercoverage & In the frozen-zero configuration, percentile coverage is 12.0\%, despite no observed incorrect conclusive decisions. & The absence of observed incorrect conclusions coexists with poor coverage of the change's magnitude.\\\bottomrule
\end{tabular}
\end{table}
\FloatBarrier

\section{Scope of the descriptive Census comparison}
\label{app:census}
The replay uses 2013 for training and ten evaluation waves in 2014--2019 and 2021--2024. It omits 2020, when standard ACS one-year estimates were not released and the public-use microdata instead used experimental weights \citep{census2020Quality,censusComparingACS}. The observed-wave sequence therefore includes a two-year calendar step.

The replay adapts ACSEmployment \citep{ding2021retiring} using ACS Public Use Microdata \citep{census2024pums}. The target is ESR=1, civilian employed at work, versus every other retained response; it is not an employed-versus-unemployed restriction. Records satisfy $16<\mathrm{AGEP}<90$ and $\mathrm{PWGTP}\ge1$. The race analysis restricts fitting and evaluation to White-alone and Black-alone categories, without an ethnicity exclusion. These administrative categories and the binary comparison do not represent all dimensions of group identity.

The 2013 feature map standardises age and uses category indicators for educational attainment, marital status, harmonised relationship, disability, employment status of parents, citizenship, mobility, military service, ancestry, nativity, and hearing, vision and cognitive difficulty. Sex and race are excluded as predictors. Missing values receive an explicit category; unseen future categories do not expand the training vocabulary. Logistic fitting and the prediction cutoff match the simulation settings. A state-wave exceeding 25,000 filtered records is sampled without replacement before group restriction, using seed $20260901+100j+t$ for FIPS code $j$ and window $t=-1,0,\ldots,9$.

A state is eligible only if every retained training and evaluation wave contains at least 500 usable records, at least 100 per group and both outcome classes, with nonzero denominators for every group--outcome stratum. This retrospective criterion uses future waves. The original 30\% calibration split uses seed 20260901 and disjoint states. For the race comparison, calibration FIPS codes are $\{5,8,12,20,25,28,34,36,37,45,47,53\}$ and evaluation codes are $\{1,4,6,9,10,13,15,17,18,19,21,22,24,26,27,29,31,32,35,39,40,41,42,44,48,49,51,54,55\}$.

Monitoring uses the first deployment wave as reference, pooled calibration-state deviations for stream scales, allowance multiplier 0.5 and sixteen geometrically spaced threshold candidates from 0.1 to 12. Full replays select the threshold whose mean refit count is closest to three, with ties favouring the larger threshold. Calibration uses evaluation-era waves and is therefore retrospective. The highlighted cadence example follows a fixed schedule independently of these monitoring thresholds.

The weighting comparison summarises the eligible records after capping. Both equal-record and person-weighted summaries calculate rates within each state, wave, group and outcome stratum, then form the disparity contrast and average it equally over the ten observed waves and then across states. Person weights alter these rates while predictions and update actions remain fixed. These finite-record summaries assess how weighting changes representation and the estimand. They do not estimate national quantities, survey-design variance, interval coverage or causal policy effects.

\section{Interval form and resample size}
\label{app:history-comparator}
\label{app:basic}

The initial saved-output comparison reconstructs 58,000 full-size basic intervals across seventeen controlled configurations and the subgroup-drift and combined-drift cadence cells, each containing twenty histories, both rates and all three audit sizes. It retains 2,000 audits per controlled configuration and 100 repeats per history--rate--size unit. Full-size percentile and basic intervals share bootstrap draws and width. Full-size versus reduced-size basic uses the same audits and interval form with different saved draws, retaining the primary exponent of 0.75.

For each method, a history audit is excluded from coverage calculations if either interval endpoint lies within $10^{-5}$ pp of its population reference. Paired differences use audits eligible for both methods. The single unresolved original-percentile coverage case remains excluded and all decision references remain resolved. At the smallest TPR size and the 0.5 pp margin, eight of the twenty subgroup-drift reduced-size basic false reductions concern a true increase beyond 0.5 pp and twelve lie within the margin. The combined-drift counts are two and thirteen out of fifteen. All four margins are evaluated with directional counts and eligible denominators, both error denominators, widths and paired Monte Carlo standard errors.

Figure~\ref{fig:history-comparator} uses positions 1--20 in fixed ID order: 2, 31, 48, 122, 128, 131, 159, 162, 184, 207, 208, 232, 247, 259, 270, 291, 325, 326, 331, 363.

Figure~\ref{fig:history-comparator} reports the full-size basic minus full-size percentile and reduced-size basic minus full-size basic contrasts, with paired Monte Carlo standard errors. Each contrast uses 100 aligned audits per history with resolved population references.

The direct comparison of reduced-size basic with percentile intervals uses all twenty originally selected subgroup-drift cadence histories in the positive-signal interaction cell, both rates and all three audit sizes, with $\gamma=0.75$. All 120 history--rate--size comparisons have 100 aligned audits and no numerically unresolved coverage or decision references.

At increasing audit sizes, reduced-size basic intervals have lower TPR coverage than full-size percentile intervals in \HistoryTPRLowerSmall, \HistoryTPRLowerMedium{} and \HistoryTPRLowerLarge{} histories, with median changes of $\HistoryTPRMedianSmall$, $\HistoryTPRMedianMedium$ and $\HistoryTPRMedianLarge$ pp. The mean frequency of correct conclusive decisions is higher under reduced-size basic intervals at all three sizes.

Mean posterior TPR coverage across the twenty histories ranges from 91.3\% to 93.3\% across audit sizes. History 259 has posterior coverage of 58/100, 60/100 and 74/100 and was highlighted retrospectively for these low values, so its pointwise Monte Carlo intervals do not adjust for that selection. For comparison, its full-size percentile coverage is \BootstrapHistoryCoverageSmall/100, \BootstrapHistoryCoverageMedium/100 and \BootstrapHistoryCoverageLarge/100, while reduced-size basic coverage is \MOutHistoryCoverageSmall/100, \MOutHistoryCoverageMedium/100 and \MOutHistoryCoverageLarge/100 at the same increasing audit sizes.

As audit size increases in this cell, the proportion of correct conclusive full-size percentile decisions rises from 3.9\% to 56.6\% for TPR and from 67.2\% to 85.5\% for FPR. Projection covers every saved target but leaves every audit of a history with model changes inconclusive at the smallest size. At the largest size, its mean inconclusive frequency remains at least 40.0\% in every nonempty comparison involving such histories.

For event $E$ (coverage or an incorrect conclusive decision), let $Z_b=1\{E_{M,b}\}-1\{E_{P,b}\}$ for audit repeat $b$. The reported difference is $100\overline Z$ pp and its paired Monte Carlo standard error is $100s_Z/\sqrt{100}$, using the sample standard deviation with denominator 99. This retains the dependence between methods on the same audit. These standard errors measure Monte Carlo precision conditional on each selected history. The comparison comprises 240 history/event combinations, each with event counts and a paired standard error.

\begin{table}[htbp]\centering\small
\caption{Primary reduced-size basic minus full-size percentile. Counts give histories with lower/equal/higher frequencies, among all twenty selected histories. Means are equally weighted frequency differences in pp. Decisions use the 0.5 pp margin.}\label{tab:history-comparator}
\begin{tabular}{@{}lr rr rr@{}}\toprule
 & & \multicolumn{2}{c}{Coverage} & \multicolumn{2}{c}{Incorrect conclusive decisions}\\
Rate & $n$ & Lower/equal/higher & Mean diff. & Lower/equal/higher & Mean diff.\\\midrule
TPR & 1,250 & 18/0/2 & -13.65 & 0/8/12 & +3.05\\
TPR & 5,000 & 17/0/3 & -10.60 & 0/15/5 & +1.40\\
TPR & 20,000 & 19/0/1 & -8.25 & 0/19/1 & +0.25\\
FPR & 1,250 & 12/1/7 & -0.85 & 0/19/1 & +0.15\\
FPR & 5,000 & 6/5/9 & -0.15 & 0/19/1 & +0.05\\
FPR & 20,000 & 7/10/3 & -0.40 & 0/20/0 & +0.00\\
\bottomrule\end{tabular}
\end{table}

\begingroup
\captionsetup{textformat=finalcaptionmoutnpanel}
\begin{table}[htbp]\centering\small
\caption{Cadence in the selected positive-signal interaction cell under subgroup drift: equal means across all 20 retained histories, each with 100 audits. M denotes the reduced-size basic interval with $\gamma=0.75$; P denotes the saved full-size percentile. Coverage and decision frequencies are percentages; decisions use the 0.5 pp margin. This cell is not representative of deployment histories.}\label{tab:moutn-panel}
\begin{tabular}{@{}lrrrrrrr@{}}\toprule
Rate & $n$ & \multicolumn{2}{c}{Coverage (\%)} & \multicolumn{2}{c}{Correct (\%)} & \multicolumn{2}{c}{Incorrect (\%)}\\
 & & P & M & P & M & P & M \\\midrule
TPR & 1,250 & 92.20 & 78.55 & 3.90 & 25.40 & 0.15 & 3.20\\
TPR & 5,000 & 91.35 & 80.75 & 29.80 & 50.10 & 0.20 & 1.60\\
TPR & 20,000 & 93.45 & 85.20 & 56.60 & 63.65 & 0.10 & 0.35\\
FPR & 1,250 & 93.50 & 92.65 & 67.15 & 69.65 & 0.30 & 0.45\\
FPR & 5,000 & 94.75 & 94.60 & 76.35 & 76.10 & 0.00 & 0.05\\
FPR & 20,000 & 94.60 & 94.20 & 85.45 & 84.75 & 0.00 & 0.00\\
\bottomrule\end{tabular}
\par\smallskip
\begin{tabular}{@{}lrr@{}}\toprule
Same TPR audits, $n=1{,}250$ & Incorrect / all audits & Incorrect / conclusive audits\\\midrule
Full-size percentile & 3/2,000 (0.15\%) & 3/81 (3.70\%)\\
Reduced-size basic & 64/2,000 (3.20\%) & 64/572 (11.19\%)\\
\bottomrule\end{tabular}

\end{table}

\endgroup

Table~\ref{tab:reporting}'s conclusive projection example is history 2, FPR, repeat 0, with 20,000 observations per window and seven windows with nonidentical prediction functions. The estimate is $-8.508$ pp, the saved population reference is $-8.402$ pp, and the interval $[\ResolvedProjectionLower,\ResolvedProjectionUpper]$ pp lies wholly below $-0.5$ pp. This retrospectively selected example illustrates a conclusive outcome, while Appendix~\ref{app:boundary-cases} gives the construction's coverage justification. Estimated FPR changes are $+0.231$ pp for group 0 and $-8.277$ pp for group 1, so the point estimates show an FPR increase for group 0 alongside the disparity reduction. These component estimates require separate uncertainty statements.

\FloatBarrier
\section{Internal robustness across drift settings}
\label{app:replication}
This internal robustness check replaces subgroup-only drift with combined drift while retaining positive signal, the interaction representation and cadence P2. It uses all twenty histories originally hash-selected in that cell, both rates, all three nested audit sizes and all 100 repeats. The cell and reporting scope were fixed after the earlier findings and parent-panel analyses, but before the new comparator results. Histories and exponents were retained without screening on the new method's performance. Both cells use the same simulator and calibration procedure.

An audit-seed check covers forty evaluation histories across the two cells. Evaluation seeds include drift regime and replication ID, and audit seeds additionally include the full cell. The two sets of twenty evaluation seeds and two sets of 20,000 audit streams are disjoint. Each audit stream supplies one window of a repeat, with nested prefixes for the three sizes.

The interval implementation is unchanged. The primary exponent is 0.75, sensitivities are 0.65 and 0.85, and all four margins remain fixed. A separate namespace determines the new interval-root seeds; the same prefix-stable stratum routine supplies 10,000 draws and 360 prespecified 40,000-draw checks, one audit per history--rate--size unit at every exponent. The calculation produces 36,000 intervals at the 10,000-draw budget across the three exponents. It uses the existing observations, fitted functions, update actions, audit samples and population references, while original method endpoints remain unchanged.

For each paired event difference we use the formula in Appendix~\ref{app:history-comparator}, restricted to the intersection of the two methods' numerically evaluable audits. Both marginal denominators and the common denominator use the original $10^{-5}$ pp tolerance. Decision comparisons exclude a numerically unresolved true margin category. Comparisons across rates and nested sizes are dependent, and ratios with zero denominators are undefined. Every original panel comparator and fixed exponent is evaluated at all four margins for all twenty combined-drift histories, with both error denominators retained at the primary margin.

One saved bootstrap TPR coverage repeat for history 14 at 20,000 observations per window is numerically unresolved; that history uses 99 repeats for P coverage and the paired difference. The separately reported M coverage retains its 100 evaluable repeats. All other method--history--size coverage series and all decision references are resolved. None of the new intervals is flagged as empirically degenerate or insufficient. The 360 higher-draw checks have a maximum endpoint shift of 0.0961 pp. One $\gamma=0.85$ check changes an inconclusive decision to a correct increase conclusion, with no other transitions and no erroneous conclusions before or after among those checks. The higher-draw checks assess numerical sensitivity for these audits.

\begingroup
\captionsetup{textformat=finalcaptionreplicationmain}
\begin{table}[htbp]\centering\small
\caption{Non-confirmatory internal robustness check with disjoint seeds: twenty combined-drift cadence histories, 100 audits each. P is the saved full-size percentile; M uses fixed $\gamma=0.75$. Frequencies are equal history means (\%); the margin is 0.5 pp. The final column counts lower/equal/higher M coverage on paired evaluable audits. One P TPR history at $n=20{,}000$ has 99 evaluable coverage repeats; its paired comparison also uses 99. Other coverage series retain 100.}\label{tab:replication}
\begin{tabular}{@{}lr rr rr rr r@{}}\toprule
 & & \multicolumn{2}{c}{Coverage} & \multicolumn{2}{c}{Correct} & \multicolumn{2}{c}{Incorrect} & Coverage histories\\
Rate & $n$ & P & M & P & M & P & M & Lower/equal/higher\\\midrule
TPR & 1,250 & 94.50 & 84.45 & 15.85 & 29.10 & 0.10 & 2.05 & 18/0/2\\
TPR & 5,000 & 94.45 & 85.75 & 36.80 & 44.45 & 0.15 & 1.40 & 19/0/1\\
TPR & 20,000 & 95.25 & 88.25 & 55.05 & 59.35 & 0.20 & 0.70 & 18/2/0\\
FPR & 1,250 & 91.10 & 86.55 & 35.25 & 49.75 & 0.00 & 0.70 & 13/0/7\\
FPR & 5,000 & 92.00 & 90.00 & 57.25 & 62.40 & 0.00 & 0.00 & 15/1/4\\
FPR & 20,000 & 93.35 & 89.60 & 69.30 & 73.45 & 0.15 & 0.40 & 15/1/4\\
\bottomrule\end{tabular}
\end{table}

\endgroup

\FloatBarrier
\FloatBarrier

\section{Paired decision transitions from saved endpoints}
\label{app:transitions}
The transition analysis follows conclusions for the same audits across interval procedures. It covers seventeen controlled configurations with 2,000 audits each, plus twenty subgroup-drift and twenty combined-drift cadence histories, both rates, all three sizes and 100 repeats per history--rate--size unit. All four fixed margins are reported, with 0.5 pp primary. Reduced-size basic retains $\gamma=0.75$ and the earlier 0.65 and 0.85 sensitivities. The analysis uses saved endpoints, and Appendix~\ref{app:evidence} records its chronology.

\begingroup
\captionsetup{textformat=finalcaptiontransitionsmain}
\begin{table}[htbp]\centering\small
\caption{Paired decision transitions in TPR at $n=1{,}250$, margin 0.5 pp. Each block contains the same 2,000 audits of twenty histories. Rows are the earlier method, columns the later method: I = inconclusive, C = correct conclusive, W = incorrect conclusive. All nine cells, including lost conclusions and errors that persist, are retained. These are descriptive pooled counts, not a binomial validity assessment.}\label{tab:transitions-main}
\begin{tabular}{@{}lllrrr@{}}\toprule
Drift & Pair & From & To I & To C & To W\\\midrule
Subgroup & P$\to$B & I & 1580 & 299 & 40\\
Subgroup & P$\to$B & C & 1 & 77 & 0\\
Subgroup & P$\to$B & W & 0 & 0 & 3\\
\midrule
Subgroup & B$\to$M & I & 1428 & 132 & 21\\
Subgroup & B$\to$M & C & 0 & 376 & 0\\
Subgroup & B$\to$M & W & 0 & 0 & 43\\
\midrule
Combined & P$\to$B & I & 1466 & 189 & 26\\
Combined & P$\to$B & C & 5 & 312 & 0\\
Combined & P$\to$B & W & 0 & 0 & 2\\
\midrule
Combined & B$\to$M & I & 1377 & 81 & 13\\
Combined & B$\to$M & C & 0 & 501 & 0\\
Combined & B$\to$M & W & 0 & 0 & 28\\
\bottomrule\end{tabular}
\end{table}

\endgroup

For each of P$\to$B, B$\to$M and P$\to$M, alignment requires identical source, history/configuration, policy, rate, size and repeat IDs, point estimates and the original identity/insufficiency flags. Method-specific empirical-degeneracy flags are retained. Each unit supplies a full $4\times4$ matrix of reduction, increase, within-margin and inconclusive categories, retaining the true category and each method's false-declaration direction. Its $3\times3$ status matrix distinguishes inconclusive, correct conclusive and incorrect conclusive outputs. Zero cells, lost conclusions and switches between conclusive categories are retained. At the primary margin, none of these audits switches directly between different conclusive categories.

Transitions use the intersection of decision-eligible audits. There are no unresolved true categories and no decision exclusions at any of the four margins in this scope. Identity and insufficient-audit rows retain their original analytic intervals and stay in the decision denominator; empirical degeneracy does not establish identity. Coverage eligibility remains separate: history 14, combined-drift TPR at $n=20{,}000$, excludes repeat 83 for P coverage, retaining 99 coverage-evaluable P audits and 100 B/M audits. P$\to$B and P$\to$M coverage differences therefore use 99 common audits there, while every decision transition still uses 100. All other coverage comparisons are fully reference-resolved. The original $10^{-5}$ pp tolerance and exact equality-at-margin rules are unchanged.

Every matrix row and column reconciles with the corresponding scored marginal counts. The analysis also reconciles all 3,084 method--unit--margin summaries and 9,252 paired coverage/correct/incorrect event comparisons with the previously calculated marginal results. The tabulations comprise 49,344 category cells and 27,756 status cells, with marginal and common eligible denominators, identity/insufficiency/degeneracy counts, directional eligible denominators and both total-error ratios. Wrong/all-decision-eligible and wrong/conclusive fractions remain distinct; a zero conclusive denominator yields an undefined ratio. Directional error rates instead use as their denominator audits for which the corresponding declaration would be false.

Primary-margin status transitions cover both selected cells, both rates and all three sizes, with every controlled configuration kept separate. The comparisons also include each history and all four evaluated margins. Pooled counts describe transitions within each cell, rate and size. Within each history or controlled configuration, paired event Monte Carlo standard errors use the common eligible audits and the Appendix~\ref{app:history-comparator} formula, $100s_Z/\sqrt R$. Transition cells partition these same audits, and the pooled counts are not treated as a single binomial sample.

\subsection{An additional method-disagreement illustration}
The selection rule was fixed in the local transition specification: among subgroup-drift TPR audits at $n=1{,}250$ and margin 0.5 pp, take the lexicographically first history ID and lowest repeat for which P is inconclusive and B falsely declares reduction. This error-directed retrospective selection gives history 31, repeat 93, under cadence P2. It illustrates an error and is not evidence about its prevalence or a method's validity. The original hash-preselected history 144 remains the separate worked audit in Figure~\ref{fig:worked}.

The saved estimate is $-1.901083342$ pp and population reference $+2.124278868$ pp. The saved endpoints, in pp, are
\[
\begin{array}{c rr}
 & L & U\\
\mathrm{P} & -3.202933913 & 1.594105794\\
\mathrm{B} & -5.396272477 & -0.599232770\\
\mathrm{M} & -4.884772902 & -0.596661153
\end{array}
\]
In the saved 10,000-draw results, full-size percentile intervals exclude the population reference but remain inconclusive. Reflecting the percentile interval shifts both endpoints by $2\widehat D-L_P-U_P=-2.193338564$ pp and moves the upper endpoint below $-0.5$ pp, producing the full-size basic interval's false reduction declaration. Reduced-size basic also declares reduction. The two full-size intervals have identical widths of 4.797039707 pp. The targeted check below leaves the full-size basic interval's limiting classification unresolved at the decision boundary, so Figure~\ref{fig:disagreement} displays the saved 10,000-draw results.

Table~\ref{tab:disagreement-counts} gives the positive-outcome paired counts. The first prediction index is updating and the second frozen. Windows 0--2 have established function identity and contribute zero analytically. Substitution of the remaining signed gaps into Equation~\eqref{eq:target}, retaining the full ten-window denominator, reconstructs $\widehat D$ to within $10^{-10}$ pp. The total audit size per window is 1,250; these counts use only positive-outcome observations in each group.

\begin{table}[htbp]\centering\small
\caption{Saved paired TPR counts for the additional disagreement illustration: history 31, repeat 93. Identity (Id.) is established from function records. The example was selected retrospectively because the saved full-size basic interval falsely declared a reduction.}\label{tab:disagreement-counts}
\begin{tabular}{@{}rrlrrrrr@{}}\toprule
Window & Group & Id. & $n_{ta}$ & $C_{00}$ & $C_{01}$ & $C_{10}$ & $C_{11}$\\\midrule
0 & 0 & Yes & 418 & 216 & 0 & 0 & 202\\
0 & 1 & Yes & 126 & 56 & 0 & 0 & 70\\
1 & 0 & Yes & 419 & 213 & 0 & 0 & 206\\
1 & 1 & Yes & 115 & 66 & 0 & 0 & 49\\
2 & 0 & Yes & 443 & 229 & 0 & 0 & 214\\
2 & 1 & Yes & 126 & 65 & 0 & 0 & 61\\
3 & 0 & No & 431 & 205 & 6 & 14 & 206\\
3 & 1 & No & 121 & 52 & 5 & 17 & 47\\
4 & 0 & No & 410 & 197 & 6 & 8 & 199\\
4 & 1 & No & 128 & 52 & 11 & 10 & 55\\
5 & 0 & No & 408 & 182 & 6 & 15 & 205\\
5 & 1 & No & 114 & 40 & 14 & 6 & 54\\
6 & 0 & No & 469 & 226 & 2 & 15 & 226\\
6 & 1 & No & 121 & 44 & 13 & 11 & 53\\
7 & 0 & No & 453 & 215 & 7 & 16 & 215\\
7 & 1 & No & 111 & 50 & 10 & 20 & 31\\
8 & 0 & No & 465 & 228 & 5 & 19 & 213\\
8 & 1 & No & 100 & 51 & 3 & 13 & 33\\
9 & 0 & No & 441 & 189 & 13 & 14 & 225\\
9 & 1 & No & 122 & 63 & 15 & 1 & 43\\\bottomrule
\end{tabular}
\end{table}

\FloatBarrier
\subsection{Targeted numerical sensitivity of Figure 1}
\label{app:figure1-stability}
The numerical check concerns the full-size basic interval's upper endpoint near $-0.5$ pp in this retrospectively selected audit. After reconciling Table~\ref{tab:disagreement-counts}, window-level function identity, the point estimate, saved endpoints and strict decision rule, we fixed the random scheme, precision criterion and maximum draw budget before generating additional draws. The check covers this audit's full-size percentile, full-size basic and reduced-size basic outputs using its existing observations, fitted functions, history and population reference. The original outputs and all aggregate summaries remain unchanged.

Both original 10,000-draw intervals reproduced to within $10^{-10}$ pp using their unchanged implementations and saved seeds. The sensitivity calculation used exactly five million draws for P/B and five million for M, with the original paired multinomial sampling and full ten-window denominator. B is the reflection of P on the same full-size draws. M retains $\gamma=0.75$, its original integer resample sizes and square-root scaling. The original full-size seed is 2489991111125334614; M's is 3362738286383530244. Increasing the original full-size implementation's draw count changes the stratum stream positions, so it does not preserve complete draw prefixes; M's separate stratum streams do preserve them. The additional draws quantify Monte Carlo sensitivity within the same audit sample.

For B, reduction requires $Q_{.025}(\widehat D^*)>c_B=2\widehat D+0.5=-3.302166684$ pp. For M, writing $E^*=\sqrt{m/N}(\widehat D_m^*-\widehat D)$, it requires $Q_{.025}(E^*)>c_M=\widehat D+0.5=-1.401083342$ pp. We assessed both strict and non-strict bootstrap CDF values at these cutoffs and each lower quantile. Two-sided exact binomial intervals for the CDFs and binomial order-statistic brackets for the quantiles each have noncoverage $0.01/6$. A union bound therefore gives at least 99\% simultaneous Monte Carlo coverage for the six assessments, conditional on the counts. The quantile brackets are conservative for atoms. Reflecting them gives brackets for the basic upper endpoints. This numerical uncertainty is distinct from audit-sampling uncertainty and the population-reference tolerance.

At five million draws, P is $[-3.300105998,1.590670842]$ pp, giving B $[-5.392837526,-0.502060686]$ pp. B's upper-endpoint bracket is $[-0.506611762,-0.497714310]$ pp, spanning $-0.5$. The count below $c_B$ is 124,451/5,000,000, with CDF interval $[0.024671712,0.025110005]$ spanning .025; the non-strict count is identical. The ten consecutive 500,000-draw block endpoints range from $-0.510630995$ to $-0.490182785$ pp and also span the boundary. The five-million-draw interval declares reduction, but its upper-endpoint uncertainty bracket crosses $-0.5$ pp, leaving the full-size basic interval's limiting classification unresolved.

M is $[-4.875535952,-0.586068742]$ pp, with upper-endpoint bracket $[-0.590127137,-0.581883955]$ pp. Its strict and non-strict CDF counts are both 103,756/5,000,000, with interval $[0.020551333,0.020952395]$ below .025. All ten block upper endpoints remain below $-0.5$ pp, ranging from $-0.592723859$ to $-0.577742293$ pp. M's false reduction is therefore separated from the boundary under this check.

Both endpoint brackets meet the prespecified half-width target of at most 0.01 pp. Only M meets the separate requirement that the entire bracket lie at least 0.01 pp from the decision boundary. We stopped at the fixed budget despite B's unresolved verdict.

\FloatBarrier
\section{Denominators for incorrect conclusive decisions}
\label{app:denominators}
For a fixed comparison, let $W$ denote an incorrect conclusive margin decision and $C$ any conclusive decision. Valid 95\% frequentist coverage bounds $\Pr(W)$ by 5\%. Since $W\subseteq C$, the error probability among conclusive audits is $\Pr(W\mid C)=\Pr(W)/\Pr(C)$ when $\Pr(C)>0$. Inconclusive audits contribute to the all-audit denominator but are excluded from the conclusive-audit denominator, so the conditional error probability need not satisfy the same 5\% bound. If no audit is conclusive, the empirical conditional ratio is undefined rather than zero.

The pooled denominator comparisons for full-size percentile and reduced-size basic intervals at $\gamma=0.75$ cover the twenty selected subgroup-drift cadence histories, both rates and all three audit sizes at the 0.5 pp margin. Here $n$ is the total number of observations per window. Counts pool 100 audits per history. Each conclusive count is the sum of correct and incorrect conclusions and agrees with Table~\ref{tab:moutn-panel}. The pooled conditional error ratio weights each history by its number of conclusive audits. These heterogeneous histories are not treated as one binomial sample, and Appendix~\ref{app:aggregate} gives the Monte Carlo precision calculations for paired contrasts between the pooled ratios.

\FloatBarrier
\section{Study chronology and numerical checks}
\label{app:evidence}
The reduced-size comparisons were specified after the original controlled and history findings and before their additional interval draws. The combined-drift reduced-size extension was specified after earlier findings and before its additional comparator results were calculated. Full-size basic reconstruction, transition summaries and aggregate precision calculations were specified before their calculation, with earlier findings known. Directional summaries and associations across histories are retrospective.

\label{app:antecedent}
The interval checks establish repeatability of selected finite-budget calculations in the tested environment. The simulator and updating policies adapt an existing study design, with the current specifications given in Appendix~\ref{app:histories}. The present study evaluates repeated-audit inference conditional on fixed histories. Its temporally averaged contrast scales the corresponding cumulative updated-minus-frozen absolute-gap contrast by $100/T$, using population rather than finite-window rates. Shared simulation and policy definitions do not establish identical trajectories, observations or outputs.

\FloatBarrier

\section{Full original-panel percentile/basic comparison}
\label{app:broad-basic}
The broader saved-output comparison retains all 64 controlled configurations with 2,000 audits each (128,000 intervals), and all 160 originally hash-selected histories, three policies, both rates, three sizes and 100 audits per unit (288,000 intervals). It therefore reconstructs 416,000 B intervals from the same number of saved P intervals, on 2,944 units. The earlier 58,000 P/B comparisons are an overlapping subset. Reduced-size basic and numerical delta are not extended to additional settings.

Full-size basic intervals are reconstructed from percentile endpoints and scored against the population contrasts.

For ordinary rows, B is exactly $[2\widehat D-U_P,2\widehat D-L_P]$. Established identity retains $[0,0]$, insufficient audits retain their full feasible intervals, and empirical degeneracy does not establish identity. Exact rational controlled targets and all original temporal weights are preserved. All four margins remain available, with 0.5 pp primary. The history-reference tolerance remains $10^{-5}$ pp: marginal coverage eligibility is method-specific, while paired coverage uses the intersection. At the primary margin, P and B each exclude two history audits from marginal coverage; all decision references are resolved. No saved endpoint or previous eligibility decision is replaced.

The analysis reconciles all 11,776 original P unit--margin summaries, all 58,000 overlapping P/B endpoint pairs, all 2,056 corresponding method summaries and all 3,084 paired event comparisons. Widths are unchanged.

There are 480 policy histories, comprising 299 with model changes and 181 with established identity throughout. For established identity, both intervals remain $[0,0]$, cover $D=0$ analytically and give within-margin decisions at every evaluated positive margin. Each cell--policy--rate--size summary averages history-specific frequencies equally. The wrong/conclusive ratio pools eligible conclusive outputs and therefore weights histories by their number of such outputs (Appendix~\ref{app:denominators}). Results describe the saved finite-draw implementations, subject to the numerical qualification in Appendix~\ref{app:figure1-stability}.

Among nonidentity history--policy comparisons, B has lower estimated TPR coverage than P in \BroadTPRLowerSmall, \BroadTPRLowerMedium{} and \BroadTPRLowerLarge{} of 299 comparisons at increasing sizes, and higher coverage in \BroadTPRHigherSmall, \BroadTPRHigherMedium{} and \BroadTPRHigherLarge{}, with ties in the remainder. The corresponding FPR counts are \BroadFPRLowerSmall, \BroadFPRLowerMedium{} and \BroadFPRLowerLarge{} lower, and \BroadFPRHigherSmall, \BroadFPRHigherMedium{} and \BroadFPRHigherLarge{} higher. These signs use integer coverage-count differences on common eligible audits, not rounded marginal percentages. These counts concern related fixed policy histories. The controlled configurations likewise show both coverage gains and losses.

\section{Decision-margin sensitivity from saved outputs}
\label{app:margins}
This retrospective analysis compares saved results at margins of 0.1, 0.25, 0.5 and 1 pp, with 0.5 pp primary. The full-panel comparison of full-size percentile and full-size basic intervals covers the 64 original controlled configurations and all original history cells, policies, rates and sizes. The comparison that also includes reduced-size basic intervals covers seventeen controlled configurations and two selected twenty-history cadence cells, with $\gamma=0.75$ for reduced-size basic. Primary numerical delta ($c=1$) is reported for the three previously highlighted controlled examples.

Each margin uses its own saved true category and original eligibility rules, including exact controlled boundaries and the $10^{-5}$ pp history-reference tolerance. As the margin increases, a fixed true contrast can move from the reduction or increase category into the within-margin category. In the selected combined-drift TPR cell, for example, the reduction/within/increase history counts change from 6/0/14 at 0.1 pp to 5/7/8 at 1 pp. Coverage counts, denominators, exclusions, widths and audit flags are unchanged across margins in every checked unit--method series. Shared audits at different margins are dependent observations, not new experiments.

In both selected TPR cells at 1,250 observations per window, moving from percentile to full-size basic intervals increases correct, incorrect and conclusive counts at every saved margin. Moving from full-size to reduced-size basic intervals does the same. Both changes also increase all-audit and conclusive-audit error fractions. For subgroup drift, full-size basic errors fall from 53 to 33 between margins of 0.1 and 1 pp, but the error fraction among conclusive audits rises from 9.20\% to 11.79\% because fewer audits are conclusive.

In frozen-zero, undercoverage coexists with zero observed incorrect conclusions at all four margins. Percentile and both basic procedures retain mixed-gap false reductions. In both-zero, percentile remains inconclusive with an undefined conclusive-audit error fraction, while both basic procedures and numerical delta give incorrect conclusive outputs.

We compare the sign of each method difference with its sign at 0.5 pp, separately for correct, incorrect and raw conclusive counts and both error fractions. Fraction signs use exact integer numerators and denominators. In the full panel excluding identity histories, the correct-count difference between full-size basic and percentile intervals reverses sign in \MarginBroadReversals{} of \MarginBroadGroups{} nonempty cell, policy, rate and audit-size groups. Identity-only groups give equal results for these two procedures across every metric and margin.

Raw conclusiveness is audits minus inconclusive outputs; reference-eligible conclusiveness is correct plus incorrect outputs. Wrong/all-decision-eligible and wrong/reference-eligible-conclusive ratios retain their separate denominators, with undefined ratios when the denominator is zero. Broad paired decision counts and restricted status-matrix marginals reconcile with the earlier results. Appendix~\ref{app:aggregate} reports Monte Carlo precision for the paired aggregate comparisons.

\clearpage
\section{Paired aggregate precision conditional on fixed histories}
\label{app:aggregate}
This analysis estimates Monte Carlo precision from saved paired summaries and joint decision-status matrices. Histories are treated as fixed strata, so the target is their specified average and uncertainty comes from repeated audits of those histories. Full-panel results for full-size percentile and full-size basic intervals are kept separate from the restricted comparison that also includes reduced-size basic intervals at $\gamma=0.75$. Controlled configurations are analysed separately. The primary margin is 0.5 pp, and results for the other three evaluated margins are included.

\subsection{Fixed weights and matched audit events}
For event $E$ and common eligible audit $i$ in history $h$, let $Z_{hi}=1\{E_{\mathrm{new},hi}\}-1\{E_{\mathrm{old},hi}\}$. With $R_h$ common audits, the saved difference and paired MCSE are $d_h=100\overline Z_h$ pp and $e_h=100s_{Z,h}/\sqrt{R_h}$, with sample-variance denominator $R_h-1$. The equally weighted fixed-history contrast and its estimated MCSE are
\[
\widehat\Delta=\sum_{h=1}^H w_hd_h,\qquad
\widehat{\mathrm{MCSE}}(\widehat\Delta)=\sqrt{\sum_{h=1}^H w_h^2e_h^2},\qquad w_h=1/H.
\]
With histories treated as fixed, the calculation includes within-history simulation variance and the covariance between procedures applied to the same audit, with no between-history sampling variance. For one controlled configuration, $H=1$, giving the saved 2,000-audit paired MCSE. The direct full-size percentile to reduced-size basic MCSE is calculated from its paired events.

The formula assumes independent audit streams across fixed histories. Audit seeds for all 160 histories follow the original hash scheme and yield 160,000 distinct history/repeat/window stream identifiers with no cross-history overlap. Policies, rates and nested sizes reuse audit information and remain separate comparisons.

Coverage uses each history's intersection of eligible audits. The report retains common and marginal denominators, exclusions, both common-subset means and the method-specific marginal difference. For combined-drift cadence TPR at $n=20{,}000$, for example, common-subset P$\to$B coverage changes by $-3.102525$ pp, whereas the marginal mean difference is $-3.100000$ pp; one P reference is unresolved. The four full-panel primary-margin exclusions are also retained individually. A history with insufficient common repeats would leave precision undefined rather than being dropped and its weight redistributed.

\subsection{Conclusive outputs and the two error ratios}
The restricted P/B/M joint status matrices retain inconclusive, correct and incorrect outputs for both methods. Every matrix reconciles with its saved marginals and paired correct/incorrect MCSEs. With no decision-reference exclusions, raw and reference-eligible conclusiveness coincide in this comparison.

For each eligible audit write $V=(W_n,C_n,W_o,C_o)$, where $W$ indicates an incorrect output and $C$ any conclusive output. Its within-history sample covariance $S_h$ retains all cross-method and numerator--denominator associations. Correct-output contrasts use $(C_n-W_n)-(C_o-W_o)$; conclusiveness uses $C_n-C_o$. Both use the fixed-history weights above.

The wrong/conclusive estimand remains the ratio of pooled wrong and conclusive counts, not the average of history-specific conditional ratios. With $a_h=R_h/\sum_jR_j$, let $\widehat\mu=\sum_h a_h\overline V_h$ and $\widehat\Sigma=\sum_h a_h^2S_h/R_h$. The saved panel has $R_h=100$, so $a_h=1/H$. Wrong/all uses $100(\widehat\mu_1-\widehat\mu_3)$; wrong/conclusive uses $100g(\widehat\mu)$, where
\[
g(\mu)=\frac{\mu_1}{\mu_2}-\frac{\mu_3}{\mu_4},\qquad
\nabla g(\mu)=\left(\frac1{\mu_2},-\frac{\mu_1}{\mu_2^2},-\frac1{\mu_4},\frac{\mu_3}{\mu_4^2}\right)^\top.
\]
The delta-method MCSE is $100\sqrt{\nabla g(\widehat\mu)^\top\widehat\Sigma\nabla g(\widehat\mu)}$. Its implementation evaluates squared within-stratum influence values to avoid cancellation when variance is exactly zero. A zero conclusive denominator leaves the ratio and its precision undefined. Outside the restricted comparison, full-panel percentile/basic results lack the saved joint status matrices needed to estimate paired covariance for contrasts between conditional-error ratios. These contrasts are therefore reported without Monte Carlo precision estimates.

\subsection{Pointwise precision and interpretation}
For positive estimated MCSE, the report gives $\widehat\Delta\pm1.959964\,\widehat{\mathrm{MCSE}}$ without truncation. These are pointwise normal or delta-method 95\% Monte Carlo approximations \citep{morris2019simulation}, not simultaneous or finite-sample coverage guarantees. Counts and denominators remain visible for sparse events and small conclusive subsets. Zero estimated MCSE is marked as no observed paired variation, with the degenerate normal interval suppressed; analytic identity is identified separately. Neither empirical zero variation nor an undefined ratio establishes zero error risk.

Paired precision is assessed for the two selected TPR cells at 1,250 observations per window. For subgroup drift, changing from full-size percentile to full-size basic intervals changes coverage by $-7.65$ pp with MCSE \AggregateSubgroupPBCoverageSE{} pp and approximate interval $[\AggregateSubgroupPBCoverageLower,\AggregateSubgroupPBCoverageUpper]$ pp. Correct conclusions increase by \AggregateSubgroupPBCorrectDelta{} pp with MCSE \AggregateSubgroupPBCorrectSE{} pp, while incorrect conclusions increase by \AggregateSubgroupPBIncorrectDelta{} pp with MCSE \AggregateSubgroupPBIncorrectSE{} pp. For the change from full-size to reduced-size basic intervals, the approximate interval for the change in the wrong/conclusive ratio spans zero at the primary margin, although point estimates are positive at every evaluated margin (Appendix~\ref{app:margins}).

Identity-only zero contrasts for percentile and full-size basic intervals follow analytically. These uncertainty summaries concern repeated audits of fixed histories and saved finite-draw implementations. They do not include future-history variation, deployment prevalence, numerical-reference uncertainty or error relative to limiting bootstrap quantiles.

\end{document}